\pdfoutput=1

\documentclass[11pt]{article}

\usepackage{ACL2023}

\usepackage{times}
\usepackage{latexsym}

\usepackage[T1]{fontenc}

\usepackage[utf8]{inputenc}

\usepackage{microtype}
\usepackage{amsmath}
\usepackage{amssymb}

\usepackage{inconsolata}
\usepackage{graphicx}
\usepackage{url}
\usepackage{soul}
\usepackage{multirow}
\usepackage{makecell}
\usepackage{arydshln}
\usepackage{todonotes}
\usepackage{subfig}
\usepackage{arydshln}
\usepackage{amsmath}
\usepackage{xcolor}
\usepackage{booktabs}
\usepackage{float}

\usepackage{gb4e}
\noautomath

\title{Linguistically Grounded Analysis of Language Models \\ using Shapley Head Values}

\author{Marcell Fekete \\
  Department of Computer Science \\
  Aalborg University \\
  Copenhagen, Denmark \\
  \texttt{mrfe@cs.aau.dk} \\\And
  Johannes Bjerva \\
  Department of Computer Science \\
  Aalborg University \\
  Copenhagen, Denmark \\
  \texttt{jbjerva@cs.aau.dk} \\}

\begin{document}
\maketitle
\begin{abstract}
Understanding how linguistic knowledge is encoded in language models is crucial for improving their generalisation capabilities. 
In this paper, we investigate the processing of morphosyntactic phenomena, by leveraging a recently proposed method for probing language models via Shapley Head Values (SHVs). 
Using the English language BLiMP dataset, we test our approach on two widely used models, BERT and RoBERTa, and compare how linguistic constructions such as anaphor agreement and filler-gap dependencies are handled.
Through quantitative pruning and qualitative clustering analysis, we demonstrate that attention heads responsible for processing related linguistic phenomena cluster together. 
Our results show that SHV-based attributions reveal distinct patterns across both models, providing insights into how language models organize and process linguistic information. 
These findings support the hypothesis that language models learn subnetworks corresponding to linguistic theory, with potential implications for cross-linguistic model analysis and interpretability in natural language processing (NLP).
\end{abstract}

\definecolor{my_turqoise}{HTML}{12DDB1}
\definecolor{my_gold}{HTML}{E5CA52}
\definecolor{my_purple}{HTML}{CB52E5}

\section{Introduction}\label{sec:introduction}

Language models gain knowledge of grammatical phenomena during pretraining.
However, exactly how this knowledge is encoded is not well understood.
While there is prior research on probing language models for morphosyntactic constructions \citep{finlayson-etal-2021-causal,mueller-etal-2022-causal,stanczak-etal-2022-neurons,acs_morphosyntactic_2023}, it is not well established if this information is crucial to the model itself, or if it is merely learned as a by-product.
We perform extensive analysis and offer evidence for a hypothesis that language models learn separate subnetworks which we can ground in linguistic theory, and are crucial to the model processing.
To our knowledge this is the first paper that uses quantitative as well as qualitative analysis grounded in linguistic theory in assessing which language model components are responsible for taking care of the processing of specific linguistic phenomena.

\begin{figure}[tb]
    \centering
    \includegraphics[width=\columnwidth]{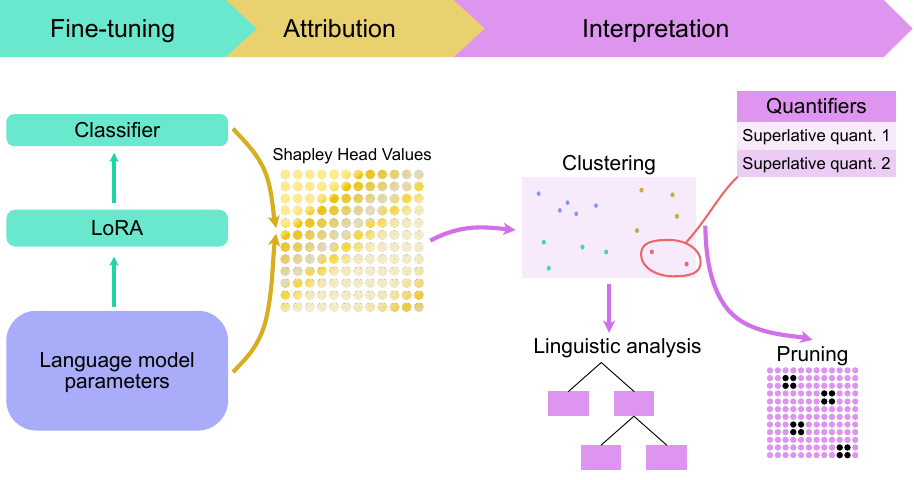}
    \caption{
    In the fine-tuning step, we train a classifier on a grammaticality judgement task.
    We carry out Shapley Head Value (SHV) attributions, and in the interpretation step, we carry out quantitative analysis using pruning as well as qualitative experiments using linguistic grounding.
    }
    \label{fig:probing-diagram}
\end{figure}

We calculate Shapley Head Values (SHVs) following the methodology of \citet{held-yang-2023-shapley}.
While they use SHVs to find attention heads that have a high contribution to certain NLP tasks such as natural language inference to improve performance, we calculate SHVs using a grammaticality classification task across 13 different phenomena and 67 different constructions from the BLiMP benchmark (\citealp{warstadt-etal-2020-blimp}, see Table \ref{tab:blimp}).
Hence, we offer a novel, linguistically grounded approach for isolating components of language models responsible for processing \emph{specific} morphosyntactic phenomena.
We cluster constructions based on SHVs, and assess the success of isolating heads responsible for processing aspects of morphosyntax using pruning based on relative importance of attention heads and linguistic analysis (see Figure \ref{fig:probing-diagram}).

In this paper, we perform an in-depth analysis using the BLiMP dataset to explore how widely used language models, such as BERT and RoBERTa, handle diverse morphosyntactic phenomena, including anaphor agreement, filler-gap dependencies, and island effects.
Through a combination of quantitative pruning and qualitative linguistic analysis, we demonstrate that certain attention heads within these models are systematically responsible for processing distinct linguistic phenomena. 
Crucially, we show that related phenomena often cluster together within the models, suggesting that language models develop internal subnetworks corresponding to theoretical linguistic categories.

We make the following contributions:

\begin{enumerate}
    \item We use Shapley Head Values (SHVs) to probe attention heads for their role in processing morphosyntactic phenomena, across two language models.
    \item We apply SHV-based clustering, showing that language models develop subnetworks corresponding to related linguistic categories.
    \item We provide qualitative linguistic analyses explaining the clustering of morphosyntactic phenomena based on SHV attributions.
    \item Quantitative pruning validates our clusters by showing localized performance impacts when relevant attention heads are removed.
    \item We release an implementation of our SHV-based probing and pruning methods.
\end{enumerate}

\section{Background}\label{sec:background}

\begin{table*}[h]
    \centering
    \resizebox{\textwidth}{!}{
    \begin{tabular}{lcll}
    \toprule
    Phenomenon & $N$ & Acceptable Example & Unacceptable Example  \\
    \midrule
    \textsc{Anaphor agr.} & 2 & {Many girls insulted \textbf{themselves}.} & *{Many girls insulted \textbf{herself}.} \\

    \textsc{Arg. structure} & 7 & {Rose wasn't \textbf{disturbing} Mark.} & *{Rose wasn't \textbf{boasting} Mark.} \\

    \textsc{Binding theory} & 7 & {Carlos said that Lori helped \textbf{him}.} & *{Carlos said that Lori helped \textbf{himself}.} \\

    \textsc{Control/raising} & 5 & {There was \textbf{bound} to be a fish escaping.} & *{There was \textbf{unable} to be a fish escaping.} \\

    \textsc{Det.-noun agr.} & 8 & {Rachelle had bought that \textbf{chair}.} & *{Rachelle had bought that \textbf{chairs}.} \\

    \textsc{Ellipsis} & 2 & \makecell[l]{{Anne's doctor cleans one \textbf{important}} \\ {book and Stacey cleans a few.}} & \makecell[l]{*{Anne's doctor cleans one book} \\ {and Stacey cleans a few \textbf{important}.}} \\

    \textsc{Filler-gap dep.} & 7 & {Brett knew \textbf{what} many waiters find.} & *{Brett knew \textbf{that} many waiters find.} \\

    \textsc{Irreg. forms} & 2 & {Aaron \textbf{broke} the unicycle.} & *{Aaron \textbf{broken} the unicycle.} \\

    \textsc{Island effects} & 8 & {Whose \textbf{hat} should Tonya wear?} & *{Whose should Tonya wear \textbf{hat}?} \\

    \textsc{NPI licensing} & 7 & {The truck has \textbf{clearly} tipped over.} & *{The truck has \textbf{ever} tipped over.} \\

    \textsc{Quantifiers} & 4 & {No boy knew \textbf{fewer than} six guys.} & *{No boy knew \textbf{at most} six guys.} \\

    \textsc{S-selection} & 2 & {\textbf{Carrie} isn't listening to Jodi.} & *{\textbf{That movie theater} isn't listening to Jodi.} \\

    \textsc{Subj.-verb agr.} & 6 & {These casseroles \textbf{disgust} Kayla.} & *{These casseroles \textbf{disgusts} Kayla.} \\

    \bottomrule
    
    \end{tabular}
    }
    \caption{Minimal pairs from the thirteen linguistic phenomena BLiMP paradigms are categorised into with the number of paradigms in each category ($N$).
    Differences are in \textbf{bold text} and, following convention, ungrammatical sentences are marked with an asterisk.
    Table adapted from \citet{warstadt-etal-2020-blimp}.}
    \label{tab:blimp}
\end{table*}

\subsection{Probing}\label{subsec:background_probing}

Previous work in attribution focuses on identifying the role that either training features or model components play with respect to various phenomena.
\textit{Training data attribution} is used to investigate, e.g., to what extent language-specific sub-networks rely on in-language examples from the training data when making predictions \citep{choenni_examining_2023}.
\citet{foroutan-etal-2022-discovering} uses \textit{iterative magnitude pruning} \citep{frankle_lottery_2019} to isolate sub-networks specific to languages and tasks like masked language modelling, named entity recognition, and natural language inference.
\textit{Structured pruning} is used for identifying attention heads important for dialogue summarisation \citep{liu_picking_2023} and cross-lingual natural language inference, e.g., with Shapley Head Values, \citep{held-yang-2023-shapley}.
\textit{Extrinsic probing} efforts such as from \citet{acs_morphosyntactic_2023} target whether morphological information is encoded by language models.
On the other hand, work in \textit{intrinsic probing} aims to reveal how exactly linguistic information is structured within a model \citep{dalvi2019one,torroba-hennigen-etal-2020-intrinsic,stanczak-etal-2022-neurons}.

Another popular technique is \textit{causal mediation analysis} (CMA), a causal method of identifying the importance of a model component to a target phenomenon \citep{pearl_direct_2001}.
Under CMA, two effects on the prediction of a model are compared: a direct effect of input intervention -- such as a text edit -- and an indirect effect.
The relative degree of the indirect effect compared to the direct effect reveals the significance of the target component to the target phenomenon.
CMA is used to investigate the role of individual neurons and attention heads in mediating gender bias \citep{vig_investigating_2020}, and to isolate neurons responsible for subject-verb agreement in English \citep{finlayson-etal-2021-causal}, and across a number of other languages \citep{mueller-etal-2022-causal}.
A limitation of CMA, however, is that text edit operations are necessarily word-level to isolate other factors such as sensitivity to specific syntactic features.
This makes it challenging to accommodate a number of morphosyntactic constructions such as island effects (see Table \ref{tab:blimp}).
Instead, we use SHVs which are particularly well-suited to probe for subnetworks responsible for processing of morphosyntactic phenomena.

Ours is certainly not the first paper to carry out the analysis of language model skills using linguistic grounding, but we believe our work is differentiated by the wider coverage in terms of morphosyntactic phenomena, as well as our efforts to localise linguistic knowledge.
Previous work explores the relation between self-attention of input tokens and dependency links \citep{htut_attention_2019,clark-etal-2019-bert}.
\citet{linzen_syntactic_2021} analyse the capabilities of LSTM and GRU models on long-distance dependencies, while \citet{wilcox_using_2024} use GPT-2 and GPT-3 to look at long-distance dependencies and island constraints.

\subsection{Shapley Head Values}\label{subsec:background_shapley}

Shapley Values originate from game theory, devised to fairly distribute a given reward among a set of players based on their relative contribution to a certain outcome \citep{shapley_17_1953,mosca-etal-2022-shap}.
They are used in model interpretability research thanks to their properties as attribution methods and satisfying theoretical properties of local accuracy, missingness and consistency
\citep{shapley_17_1953,ghorbani_neuron_2020,held-yang-2023-shapley}.
Shapley Values have the advantage over alternative, gradient-based attribution methods in that they do not need evaluation functions to be differentiable, allowing them to be applied directly.
They are also meaningfully signed with positive values reflecting positive contribution of the component and negative values reflecting the opposite \citep{held-yang-2023-shapley}..
We use Shapley Head Values (SHVs) to measure the mean marginal contributions of attention heads in a language model in a linguistically grounded scenario, approximating these values following \citet{held-yang-2023-shapley}.
Concretely, we cluster morphosyntactic phenomena based on the similarity of their associated SHVs, then analyse the resulting cluster with regards to their correspondence to linguistic theory.

\subsection{Pruning}\label{subsec:background_pruning}

According to the Lottery Ticket Hypothesis \citep{frankle_lottery_2019}, neural models contain both harmful and beneficial connections between model components with respect to a target scenario.
This means that pruning -- the removing or turning off individual neurons or attention heads -- can help us isolate subnetworks (`winning tickets') within language models \citep{pfeiffer_modular_2024}.
The common technique of pruning is to stop signals from passing through specific model components using a binary mask.
Where the goal is improving model performance, the mask will impact components that contribute negatively to the target task, language, or other use cases.
Pruning may also be utilised to discern how localised processing of a morphosyntactic phenomenon is in the model, and how generalisable this ability is to other phenomena.
In our work, pruning is thus a quantitative metric to evaluate the cohesion of SHV clusters and the success of isolating components of a subnetwork encoding the same aspects of linguistic knowledge.

\section{Methodology}\label{sec:methodology}

\subsection{Deriving SHVs}\label{subsec:method_shapley}

Following \citet{held-yang-2023-shapley}, we define SHV $\varphi_h$, for a single attention head $\textrm{Att}_h \in A$\footnote{Representing the set of all attention heads.}, to represent the mean performance improvement on the characteristic function $V$ -- as derived from performance on the evaluation metric described in Section \ref{subsubsec:grammaticality} -- if $\textrm{Att}_h$ contributes to the inference.
To be able to remove or add attention heads at will, we augment our target models with a gate $G_h = \{0, 1\}$ for each attention head.
When $G_h = 0$, the head $\textrm{Att}_h$ is removed from the inference and does not contribute to the output of the transformer.
The derivation of $\varphi_h$ requires the contribution of the head $\textrm{Att}_h$ to be measured across all $Q$ permutations of the other gates, see Equation \ref{eq:gate}.

\begin{equation}\label{eq:gate}
    \varphi_h = \frac{1}{|Q|} \sum_{A \in Q} V (A \cup h) - V(A)
\end{equation}

Calculating Equation \ref{eq:gate} for all of $N$ attention heads requires $2^N$ evaluations, which is intractable with the number of heads language models contain.
We can facilitate the computation through a number of steps \citep{ghorbani_neuron_2020,held-yang-2023-shapley}.
First, we can replace the full permutation set $Q$ in Equation \ref{eq:gate} by a randomly sampled subset of permutations via Monte Carlo simulations \citep{castro_polynomial_2009}.
Additionally, since SHV estimates are low-variance and computationally expensive to derive, we can speed up convergence by applying stopping criteria.
One of these criteria is a truncation heuristic that determines stopping the sampling of the marginal contributions of a head once $<50\%$ of attention heads remain in the permutation \citep{held-yang-2023-shapley}.

The other stopping criterion is rooted in multi-armed bandit sampling.
We want to stop sampling the marginal contributions of a head $\textrm{Att}_h$ when we reach a decrease in the variance range in the approximated $\hat{\varphi}_h$ of $\textrm{Att}_h$.
A low variance range indicates that we can be fairly confident in the degree of impact $\textrm{Att}_h$ has for the characteristic function $V$.
Our confidence interval is derived by Empirical Bernstein Bounds variance estimation \citep{maurer_empirical_2009}.
Given $t$ samples with observed variance $\sigma_t$ and a maximum variance range of $R$, the difference between the observed mean $\hat{\mu}$ and the true mean $\mu$ falls in range given by Equation $\ref{eq:ebb}$ \citep{mnih_empirical_2008} with a probability of $1 - \delta$.

\begin{equation}
    \label{eq:ebb}
    | \hat{\mu} - \mu | \leq \sigma_t \sqrt{\frac{2\operatorname{log}(3/\delta)}{t}} + \frac{3R\operatorname{log}(3/\delta)}{t}
\end{equation}

Sampling stops when the bound is less than \mbox{$|\mu - 0|$} as this identifies the SHV as positive or negative with probability $1 - \delta$, which is an efficient way to determine which heads contribute positively to the target task.
Following \citet{held-yang-2023-shapley}, we set $R=1$ and $\delta=0.1$.

\subsection{Experimental Setup}\label{subsec:method_experimental}

In the following, we introduce our experimental setup, including the dataset we use, the target task to derive SHVs, and our fine-tuning setup.

\subsubsection{BLiMP Dataset}\label{subsubsec:blimp}

The Benchmark of Linguistic Minimal Pairs (BLiMP) is a challenge set for English that aims to measure the amount of linguistic knowledge language models have on a range of morphosyntactic phenomena \citep{warstadt-etal-2020-blimp}.
Constructions are organised into 67 minimal pair paradigms each containing 1,000 sentence pairs, and the individual paradigms are organised into 13 larger categories (see Table \ref{tab:blimp}).\footnote{For a full list of paradigms, see Appendix \ref{sec:appendix_blimp}.}
While similar datasets exist for Chinese with 38 \citep{song-etal-2022-sling}, for Japanese with 39 \citep{someya-oseki-2023-jblimp}, and Russian with 45 individual paradigms \citep{taktasheva_rublimp_2024}, BLiMP is the largest in its category.

\subsubsection{Grammaticality Judgement Task}
\label{subsubsec:grammaticality}

We define the evaluation metric mentioned in Section \ref{subsec:method_shapley} for the SHV derivation as accuracy on a custom grammaticality judgement task: a classifier has to output a binary label signifying which element of a sentence pair $(s_1, s_2)$ is grammatical.
The same metric is used to derive pruning performance.
Both sentences are drawn from the same BLiMP paradigm $p$, member of the set of all paradigms $\mathcal{P}$, $p$ shuffled prior to the selection of $(s_1, s_2)$.
Shuffling is done to facilitate the classifier focusing on the underlying morphosyntactic phenomenon represented -- or violated -- in the sentences, as opposed to merely the surface features of $(s_1, s_2)$.

The performance of the classifier is measured in terms of accuracy, and it assigns label $0$ or $1$, depending on if $s_1$ or $s_2$ is grammatical:

\[ \text{BinaryClass}(s_1, s_2) = 
\begin{cases}
    0 & \text{if} \: s_1 \: \text{is grammatical,} \\
    1 & \text{if} \: s_2 \: \text{is grammatical.}
\end{cases}
\]

The advantage of this novel task formulation is that -- unlike other approaches such as simple text edits -- it can flexibly incorporate morphosyntactic phenomena where the grammatical and ungrammatical sentences differ in more than a single word (see e.g., \textsc{Ellipsis} or \textsc{Island effects} in Table \ref{tab:blimp}).

\subsubsection{Fine-Tuning Setup}\label{subsubsec:fine-tuning}

We carry out our experiments and our analysis using the monolingual English models BERT \citep{devlin-etal-2019-bert} and RoBERTa \citep{liu_roberta_2019}.
BERT has 110 million parameters and was trained on 16GB of data, while RoBERTa has 125 million parameters and was trained on 160GB.
We follow \citet{held-yang-2023-shapley} in deriving SHVs (see Section \ref{subsec:method_shapley}). However, rather than fully fine-tuning our target model weights for the target task, we use low-rank adapters (LoRA; \citealt{hu_lora_2021}) implemented with the \texttt{PEFT} library.\footnote{See \url{https://huggingface.co/docs/peft/index}.}
The main advantages of LoRA modules are their speed of training, and, more importantly, that they merely emphasise information that is already present in the original weights as opposed to reshaping the original model weights \citep{hu_lora_2021}.
This has the added advantage that it enables us to isolate and evaluate the pretrained linguistic knowledge of the models we consider.

We merge a subset of the sentence pairs from the individual BLiMP paradigms $p \in \mathcal{P}$ to create the training set for training LoRA and the classifier, and merge a smaller subset of pairs as a development set.
For each $p$, 800 pairs go to the training set, and 100 pairs are assigned to the development set.
After splitting the sentence pairs -- but before merging into the training and development sets -- we permute the set of ungrammatical sentences to avoid exact minimal pairs in order to better focus on the underlying grammatical construction (see Section \ref{subsubsec:grammaticality}).
Finally, we shuffle the order of grammatical and ungrammatical sentences to create appropriate binary training examples, and merge the selections into the training and development sets.
We retain the remaining 100 sentence pairs per paradigm to derive paradigm-specific attributions.

\subsection{Interpretation}\label{subsec:method_interpretation}

\begin{figure}
    \centering
    \includegraphics[width=\columnwidth]{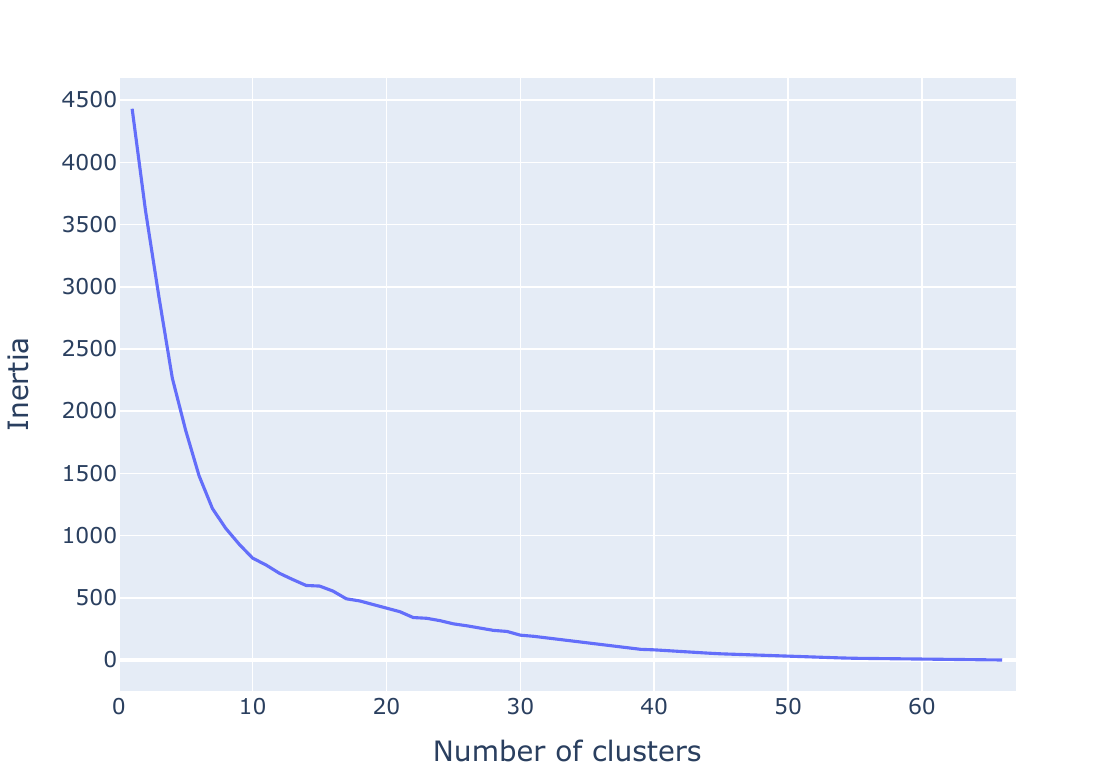}
    \caption{Clustering is done to try to optimise inertia and cluster count, resulting in an attempt at 10 clusters.}
    \label{fig:inertia}
\end{figure}

In this section, we describe the way we cluster BLiMP paradigms using SHVs, followed by how we interpret these clusters both with qualitative and quantitative means to assess how successfully we can identify subnetworks encoding specific linguistic knowledge across BERT and RoBERTa.

SHVs represent the mean marginal contributions of each attention head to the performance on the grammaticality judgement task defined in Section \ref{subsubsec:grammaticality}.
For each BLiMP paradigm $p \in \mathcal{P}$, these SHV attributions are represented as vectors $\mathbf{v}_p \in \mathbb{R}^d$ where $d$ is 144, i.e., the count of all attention heads in BERT and RoBERTa.
We scale these vectors and cluster them using $k$-means clustering, grouping paradigms together based on how similarly the individual attention heads contribute to processing the target constructions.
We decide an optimal number of clusters $k$ by calculating inertia (see Figure \ref{fig:inertia}).
Based on empirical results, we pick $k = 10$, a small enough number to also facilitate in-depth analysis.

\begin{figure}
    \centering
    \includegraphics[width=\columnwidth]{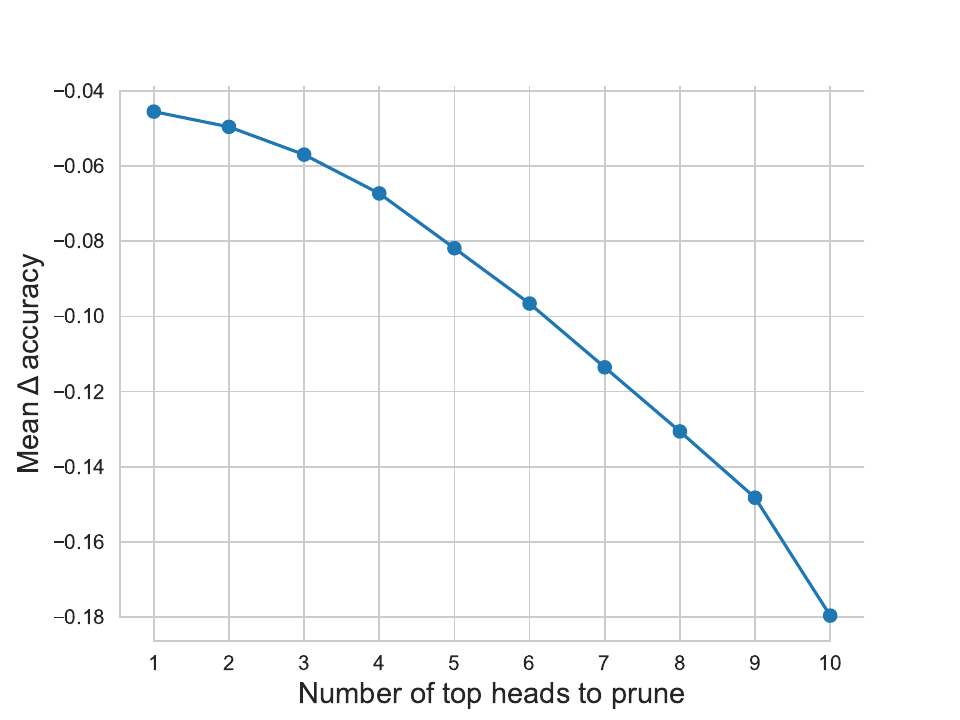}
    \caption{Mean $\Delta$ accuracy values drop in a near-linear fashion when pruning up to the top $n$ heads across paradigms
    }
    \label{fig:pruning_mean}
\end{figure}

The goal of our qualitative, linguistic analysis is to understand the potential links between diverse paradigms that are clustered together, thus validating how successfully we identified the subnetworks responsible for these constructions.
BLiMP paradigms are assigned into one of twelve categories reflecting morphosyntactic phenomena (see Table \ref{tab:blimp}).
On the highest level, our qualitative analysis is guided by the category membership of the individual paradigms -- clusters containing paradigms from the same BLiMP category are likely to be cohesive in terms of morphosyntactic phenomena they represent.
Clusters may be cohesive, however, even where category membership is not homogeneous.
Categories may represent aspects of some of the same morphosyntactic phenomena -- like how both \textsc{Binding theory} and \textsc{Anaphor agr.} are concerned with the distribution of anaphors with respect to their antecedents.
Category-level analysis can be complemented well by a sentence-level one.

Our quantitative analysis relies on pruning.
We generate binary masks for each $p$ paradigm, masking the top $n$ attention heads in terms of their SHV, then apply this mask across all other paradigms $\mathcal{P}$.
We do this to evaluate how cohesive the resulting clusters are, i.e., to what degree they represent that the language model has to apply some of the same set of morphosyntactic knowledge to process paradigms of the cluster.
It stands to reason that if the cluster is well-defined, pruning using masks within cluster should have a larger impact than when utilising these masks for paradigms that are out-of-cluster.
This is because out-of-cluster paradigms likely do not require the same set of morphosyntactic knowledge that is encoded in the relevant attention heads.
As Figure \ref{fig:pruning_mean} shows, pruning the top 10 heads already results in a large impact on accuracy across various paradigms.
Since we target only 7\% of all heads, we can observe how much of the processing of the various phenomena is localised as opposed to distributed more widely in the language model.
This way we can verify how successfully we isolated the relevant components of a given subnetwork.

\section{Results}\label{sec:results}

\begin{figure}
    \centering
    \includegraphics[width=\columnwidth]{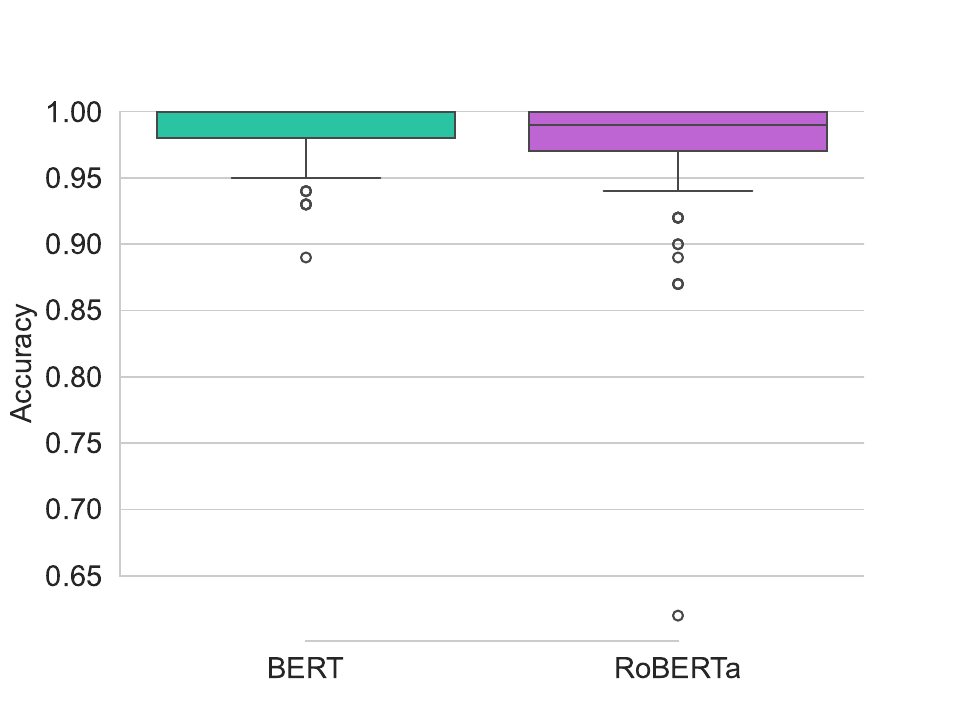}
    \caption{Distribution of baseline accuracy levels without pruning across the BERT and RoBERTa models.
    }
    \label{fig:baselines}
\end{figure}

\begin{table}[]
    \centering
    \resizebox{\columnwidth}{!}{
    \begin{tabular}{lllc}
    \toprule
     & BLiMP Paradigm  & Linguistics Term & M \\
    \midrule
    \multirow{8}{*}{
    \rotatebox{90}{\textbf{NPI Cluster}}
    } & NPI present (1) & \multirow{7}{*}{\textsc{NPI licensing}} & $\lozenge$ \\
    & NPI present (2) & & \textcolor{my_turqoise}{$\blacklozenge$} \\
    & NPI scope (`only') & & \textcolor{my_purple}{$\blacklozenge$} \\
    & NPI scope (sentential negation) & & $\lozenge$ \\
    & NPI licensor present (`only') & & $\lozenge$ \\
    & \qquad " \qquad (matrix question) & & $\lozenge$ \\
    & \qquad " \qquad (sentential negation) & & $\lozenge$ \\
    \cmidrule(rl){2-4}
    & Irregular past participle verbs & \textsc{Irregular forms} & \textcolor{my_turqoise}{$\blacklozenge$} \\
    \midrule
    \multirow{7}{*}{
    \rotatebox{90}{\textbf{Island Effects}}
    } & Adjunct island & \multirow{7}{*}{\textsc{Island effects}} & $\lozenge$ \\
    & Complex NP island & & $\lozenge$ \\
    & \makecell[l]{Coordinate structure constraint \\ (complex left branch)} & & $\lozenge$ \\
    & \qquad " \qquad (object extraction) & & \textcolor{my_purple}{$\blacklozenge$} \\
    & Left branch island (simple question) & & $\lozenge$ \\
    & Left branch island (echo question) & & $\lozenge$ \\
    & Wh-island & & \textcolor{my_purple}{$\blacklozenge$} \\
    \midrule
    \multirow{2}{*}{
    \rotatebox{90}{\makecell{\textbf{Quan-} \\ \textbf{tifiers}}}
    } & Superlative quantifiers 1 & \multirow{2}{*}{\textsc{Quantifiers}} & $\lozenge$ \\
    & Superlative quantifiers 2 & & $\lozenge$ \\
    \midrule
    \multirow{7}{*}{
    \rotatebox{90}{\textbf{Binding*}}
    } & Anaphor gender agreement & \textsc{Anaphor agr.} & \textcolor{my_purple}{$\blacklozenge$} \\
    \cmidrule(rl){2-4}
    & Animate subject trans. & \textsc{S-selection} & \textcolor{my_turqoise}{$\blacklozenge$} \\
    \cmidrule(rl){2-4}
    & Principle A (case 1) & \multirow{5}{*}{\textsc{Binding}} & \textcolor{my_turqoise}{$\blacklozenge$} \\
    & \qquad " \qquad (domain 1) & & $\lozenge$ \\
    & \qquad " \qquad \underline{(domain 2)} & & $\lozenge$ \\
    & \qquad " \qquad \underline{(domain 3)} & & $\lozenge$ \\
    & \qquad " \qquad (c-command) & & $\lozenge$ \\
    \midrule
    \multirow{11}{*}{
    \rotatebox{90}{\textbf{Filler-Gap}}
    } & Ellipsis N-bar (2) & \textsc{Ellipsis} & \textcolor{my_turqoise}{$\blacklozenge$} \\
    \cmidrule(rl){2-4}
    & \makecell[l]{Existential `there' \\ (subject raising)} & \multirow{2}{*}{\textsc{Control/raising}} & \textcolor{my_turqoise}{$\blacklozenge$} \\
    & Tough vs raising (2) & & \textcolor{my_purple}{$\blacklozenge$} \\
    \cmidrule(rl){2-4}
    & Principle A (case 1) & \multirow{3}{*}{\textsc{Binding}} & \textcolor{my_purple}{$\blacklozenge$} \\
    & \qquad " \qquad (case 2) & & $\lozenge$ \\
    & \qquad " \qquad (reconstruction) & & $\lozenge$ \\
    \cmidrule(rl){2-4}
    & Wh-questions (object gap) & \multirow{5}{*}{\textsc{Filler-gap dep.}} & $\lozenge$ \\
    & \qquad " \qquad (subject gap) & & $\lozenge$ \\
    & \makecell[l]{\qquad " \qquad \\ (subject gap, long distance)} & & $\lozenge$ \\
    & Wh vs `that' (no gap) & & $\lozenge$ \\
    & \qquad " \qquad (no gap, long distance) & & $\lozenge$ \\
    \midrule
    \multirow{5}{*}{
    \rotatebox{90}{\textbf{Wh vs `That'}}
    } & Inchoative & \multirow{2}{*}{\textsc{Arg. structure}} & \textcolor{my_purple}{$\blacklozenge$} \\
    & Intransitive & & \textcolor{my_purple}{$\blacklozenge$} \\
    \cmidrule(rl){2-4}
    & Tough vs raising (1) & \textsc{Control/raising} & \textcolor{my_purple}{$\blacklozenge$} \\
    \cmidrule(rl){2-4}
    & Wh vs `that' (with gap) & \multirow{2}{*}{\textsc{Filler-gap dep.}} & $\lozenge$ \\
    & \qquad " \qquad (with gap, long distance) & & $\lozenge$ \\
    \bottomrule
    \end{tabular}
    }
    \caption{
    Clusters that emerge in both models with shared paradigms marked with $\lozenge$.
    A subset of paradigms only appear in in BERT (\textcolor{my_turqoise}{$\blacklozenge$}) or RoBERTa (\textcolor{my_purple}{$\blacklozenge$}).
    Clusters are named by the authors based on majority membership.\footnotemark
    }
    \label{tab:matching_clusters}
\end{table}

\footnotetext{There are two different \textbf{Binding} clusters in BERT, the paradigms in the smaller cluster are \underline{underlined}.}

\begin{table}[]
    \centering
    \resizebox{0.95\columnwidth}{!}{
    \begin{tabular}{lllc}
    \toprule
     & BLiMP Paradigm & Linguistics Term & M \\
    \midrule
    \multirow{12}{*}{
    \rotatebox{90}{\textbf{Det.-Noun Cluster}}
    } & Determiner-noun agr. (1) & \multirow{8}{*}{\textsc{Det.-noun agr.}} & \textcolor{my_purple}{$\blacklozenge$} \\
    & \qquad " \qquad (2) & & \textcolor{my_purple}{$\blacklozenge$} \\
    & \qquad " \qquad (irregular, 1) & & \textcolor{my_purple}{$\blacklozenge$} \\
    & \qquad " \qquad (irregular, 2) & & \textcolor{my_purple}{$\blacklozenge$} \\
    & \qquad " \qquad (with adjective, 1) & & \textcolor{my_purple}{$\blacklozenge$} \\
    & \qquad " \qquad (with adjective, 2) & & \textcolor{my_purple}{$\blacklozenge$} \\
    & \makecell[l]{\qquad " \qquad  (with adjective, \\ \qquad\qquad irregular 1)} & & \textcolor{my_purple}{$\blacklozenge$} \\
    & \makecell[l]{\qquad " \qquad (with adjective, \\ \qquad\qquad irregular 2)} & & \textcolor{my_purple}{$\blacklozenge$} \\
    \cmidrule(rl){2-4}
    & \makecell[l]{Distractor agreement \\ (relational noun)} & \multirow{3}{*}{\textsc{Subject-verb agr.}} & \textcolor{my_purple}{$\blacklozenge$} \\
    & Regular plural subject-verb agr. (1) & & \textcolor{my_purple}{$\blacklozenge$} \\
    & Regular plural subject-verb agr. (2) & & \textcolor{my_purple}{$\blacklozenge$} \\
    \cmidrule(rl){2-4}
    & Transitive & \textsc{Arg. structure} & \textcolor{my_purple}{$\blacklozenge$} \\
    \bottomrule
    \end{tabular}
    }
    \caption{
    Paradigms in the \textbf{Det.-Noun Cluster} in RoBERTa (\textcolor{my_purple}{$\blacklozenge$}).
    }
    \label{tab:only_roberta_cluster}
\end{table}

We cluster the BLiMP paradigms into 10 individual clusters using the $k$-means algorithm based on the SHV vectors across BERT and RoBERTa.
While clustering is stochastic, independent runs of the algorithm do not seem to differ significantly from one another.\footnote{See Appendix \ref{sec:appendix_purity} for analysis.}
As Table \ref{tab:matching_clusters} demonstrates, six of the ten clusters show significant correspondence across the two language models.
Some clusters, however, are less generalisable across the two models.
The RoBERTa cluster in Table \ref{tab:only_roberta_cluster} contains paradigms representing a number of diverse linguistic phenomena, and the BERT cluster featuring \textsc{Det.-noun agreement} paradigms contains an even wider range of constructions.\footnote{See all clusters in Appendix \ref{sec:appendix_clusters}.}

\begin{figure}
    \centering
    \includegraphics[width=\linewidth]{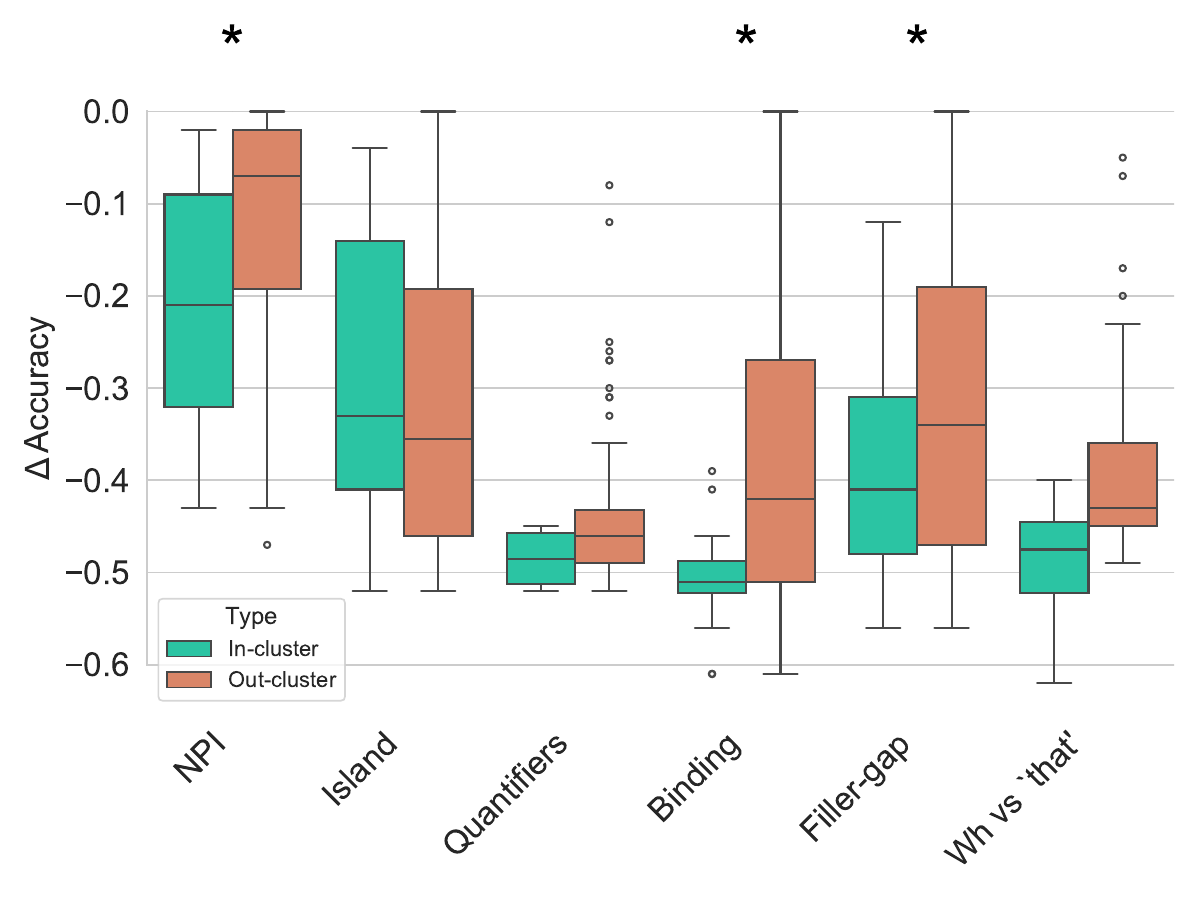}
    \caption{
    Impact in terms of $\Delta$ in accuracy in-cluster versus out-of-cluster across six BERT clusters when pruning the top 10 attention heads.
    Asterisks (*) show where the difference in distribution of the delta values is significant at $\alpha\leq0.001$ after applying Bonferroni correction \citep{dunn_multiple_1961} (see Appendix \ref{sec:appendix_ttest}).
    }
    \label{fig:bert_in_out_cluster}
\end{figure}

Figure \ref{fig:bert_in_out_cluster} compares the relative impact of pruning in- and out-of-cluster across the BERT clusters in Table \ref{tab:matching_clusters}, i.e., the ones that have related RoBERTa clusters.
This impact is measured in the change ($\Delta$) from the baseline values in terms of accuracy on the grammaticality judgement task when pruning the top 10 attention heads.
When in-cluster pruning in general results in a more dramatic negative $\Delta$ than pruning out-of-cluster, we can surmise that the top 10 attention heads contribute significantly to the processing of the target paradigms, thus we successfully identified the most important parts of a subnetwork.
In other cases, e.g., the \textbf{Island} or \textbf{Quantifiers} cluster, either the target phenomenon is not localised enough for the top 10 heads to cause a significant impact, or the clusters are not homogeneous or exclusive enough with respect to the phenomena they contain.

\section{Discussion}\label{sec:discussion}

In the following, we discuss our most important findings regarding the clusters created via SHV attributions.
The membership of these clusters indicates which morphosyntactic phenomena are processed using the same subnetworks by BERT and RoBERTa, thus showing how the models generalise over these constructions.

\paragraph{Consistency across models}
Cluster membership across the BERT and RoBERTa models largely corresponds between the majority -- 6 out of 10 -- of the clusters (see Table \ref{tab:matching_clusters}).

\paragraph{Successful grouping of linguistic categories}
In many cases, BLiMP paradigms from the same linguistic categories appear in their own clusters.
This is especially true for the clusters that are consistent across the language models.
For instance, \textsc{NPI licensing} or \textsc{Binding} paradigms mostly appear in their \textbf{NPI} or \textbf{Binding} cluster, respectively (see Table \ref{tab:matching_clusters}).
\mbox{\textsc{Det.-noun agr.}} paradigms are in their own -- though less homogeneous -- cluster in the RoBERTa model (Table \ref{tab:only_roberta_cluster}) and the BERT model as well (see Table \ref{tab:bert_roberta_clusters_2} in Appendix \ref{sec:appendix_clusters}).

\paragraph{Common ground across linguistic categories}
While clusters do tend to collect BLiMP paradigms from the same categories, exceptions, i.e., the presence of paradigms from other linguistic categories, can often be explained by linguistic analysis.
Take the \textbf{Filler-gap} cluster (Table \ref{tab:matching_clusters}).
Paradigms in \textsc{Filler-gap dep.} typically represent the fronting of \textcolor{purple}{linguistic material} that can be analysed to have \textcolor{blue}{originated} in a different clause (\ref{ex:filler-gap}).
Similarly, relevant \textsc{Control/raising} (\ref{ex:control-raising}) and \textsc{Binding} (\ref{ex:binding_filler}) paradigms also deal with the licensing of raised material.
Finally, \textsc{Ellipsis} can be analysed as a link between an antecedent and consequent clause which allows the omission of linguistic material ($e$ in \ref{ex:ellipsis}).\footnote{The symbol \textcolor{blue}{$t$} for trace is used as a convention to indicate where the raised material originates from, while \textcolor{blue}{$e$} represents the omitted material.}

\begin{exe}
    \ex\label{ex:filler-gap} Wayne has revealed \textcolor{purple}{who}/*that most hospitals admired \textcolor{blue}{$t$}.
    \ex\label{ex:control-raising} \textcolor{purple}{There} was bound/*unable to be a \textcolor{blue}{fish} escaping.
    \ex\label{ex:binding_filler} It's \textcolor{purple}{himself} that this cashier attacked \textcolor{blue}{$t$} /*It's himself that attacked this cashier.
    \ex\label{ex:ellipsis} This print scares a lot \textcolor{purple}{of busy senators} and Benjamin scares a few \textcolor{blue}{$e$} /*This print scares a lot of senators and Benjamin scares a few busy.
\end{exe}

The \textbf{Binding} cluster contains BLiMP paradigms related to phenomena concerning anaphors.
\textsc{Binding} paradigms represent the various licensing restrictions of anaphors which is often represented by the presence or lack of gender agreement between the \textcolor{purple}{anaphor} and the \textcolor{blue}{antecedent noun} (\ref{ex:binding}), similarly to the relevant \textsc{Anaphor agr.} paradigm (\ref{ex:gender}).

\begin{exe}
    \ex\label{ex:binding} Gina explains \textcolor{blue}{Alan} fires \textcolor{purple}{himself}/*Alan explains Gina fires himself.
    \ex\label{ex:gender} A \textcolor{blue}{girl} couldn't reveal \textcolor{purple}{herself}/*himself.
\end{exe}

Finally, the \textsc{Det.-noun agr.} and \textsc{Subject-verb agr.} paradigms in the RoBERTa cluster in Table \ref{tab:only_roberta_cluster} are connected by the fact that both phenomena concern number agreement (\ref{ex:plural1} and \ref{ex:plural2}).

\begin{exe}
    \ex
    \begin{xlist}
        \ex\label{ex:plural1} Raymond is selling this sketch/*sketches.
        \ex\label{ex:plural2} The students/*student perform.
    \end{xlist}
\end{exe}

\paragraph{Linguistic knowledge is often strongly localised}
The difference between in-cluster and out-of-cluster pruning impact on $\Delta$ accuracy indicates we identified the majority of attention heads representing the full construction-specific subnetwork (see Figure \ref{fig:bert_in_out_cluster}).
In a subset of cases this is not the case: either the subnetwork is larger, or the relevant knowledge is more widely distributed across model components.
Three out of the six clusters show a significant difference between in-cluster and out-of-cluster $\Delta$ accuracy.
This is very unlikely to occur randomly: in our experiments with 125 random clusters of varying sizes, we found it to happen only in two cases.

\paragraph{Different degrees of sensitivity between models}
SHVs prove useful to cluster related BLiMP paradigms into clusters, but it is clear that the RoBERTa model is somewhat more discerning with regards to morphosyntactic phenomena.
In the case of BERT, 25 of all 67 paradigms are assigned to a single cluster representing 9 different linguistic categories (see \textbf{BERT 3} in Table \ref{tab:bert_roberta_clusters_2}, Appendix \ref{sec:appendix_clusters}).
RoBERTa has no such cluster that collects this number of unrelated linguistics constructions.
The discrepancy between the two language models may lie in the comparably higher degree of linguistic knowledge obtained by RoBERTa thanks to the fact that it is pretrained on vastly more data than BERT for more training steps, and that it has more model parameters (see Section \ref{subsubsec:fine-tuning}).
This is also shown how its performance surpasses BERT on many metrics \citep{liu_roberta_2019}.

\noindent
\linebreak
These points show that language models encode \emph{at least} a subset of all morphosyntactic knowledge in subnetworks that our methodology can identify.
This enables us to evaluate the generalisation abilities of language models between related linguistic constructions.
We can additionally find that the success of generalisation is impacted by the depth of pretraining, i.e., the size of pretraining data and the number of training steps.

\section{Conclusion}

We apply intrinsic probing on two commonly used language models, BERT and RoBERTa, in order to investigate how linguistic knowledge is represented and organised internally.
We show that attributions based on SHVs can be used to identify attention heads of subnetworks that generalise across related morphosyntactic phenomena, and allow us to carry out a linguistically grounded analysis.
In this, we showcase that SHVs are well-suited to provide linguistically interpretable insights into the inner workings of language models, beyond the task-specific investigations carried out by \citet{held-yang-2023-shapley}.
Additionally, our pruning analysis demonstrates that in many cases, the identified attention heads are crucial components of the identified subnetworks;
we find that switching these attention heads off severely impacts the grammaticality judgement of the language models.
In future work, our methods can prove valuable in describing subnetworks specific to linguistic knowledge in language models.
This might be particularly valuable in cross-lingual settings.

\section*{Limitations}

A major limitation for our work is the difficulty of scaling it to new languages and to new language models.
First, the coverage of datasets of linguistic minimal pairs for languages other than English lags behind that of BLiMP.
Next, linguistically grounded clustering analysis requires an in-depth understanding of the specific morphosyntactic phenomena in a language, as well as considerable manual effort.
This effort is a bottleneck on how readily the analysis can be carried out for other language models and, of course, other languages.
Additionally, sentences in the BLiMP dataset are generated from expert templates using a predefined vocabulary, which means they tend to be semantically empty as well as heavily formulaic.
However, since this is true for both grammatical and ungrammatical sentences, this fact is not likely to cause problems for our attributions.
Furthermore, the derivation of SHVs requires the use of sampling techniques and truncation heuristics to make computation tractable.
These techniques introduce uncertainty, which means there is a slight chance they might influence the accuracy of the attribution process.
Moreover, the binary gates used to zero attention head activations may harm model performance since these zero-values are out-of-distribution for the language model.
This means the actual importance of attention heads somewhat difficult to discombobulate from the impact of the zero ablations.
Finally, we are carrying out attributions on the level of the attention head, rather than on the level of the neuron.
This improves tractability and facilitates the qualitative analysis, but it may result in a loss of granularity.
In future work, it might be worth exploring what we can gain from neuron-level attributions.
\section*{Ethics Statement}

This paper deals with a linguistically grounded analysis of the inner workings of language models.
Our work adheres to the ACL ethics guidelines.

\section*{Acknowledgements}
MF and JB are funded by the Carlsberg Foundation, under the \textit{Semper Ardens: Accelerate} programme (project nr. CF21-0454). 

\bibliography{anthology,custom,more}

\begin{thebibliography}{38}
\expandafter\ifx\csname natexlab\endcsname\relax\def\natexlab#1{#1}\fi

\bibitem[{\'{A}cs et~al.(2023)\'{A}cs, Hamerlik, Schwartz, Smith, and Kornai}]{acs_morphosyntactic_2023}
Judit \'{A}cs, Endre Hamerlik, Roy Schwartz, Noah~A. Smith, and Andras Kornai. 2023.
\newblock \href {https://doi.org/10.1017/S1351324923000190} {Morphosyntactic probing of multilingual {BERT} models}.
\newblock \emph{Natural Language Engineering}, pages 1--40.

\bibitem[{Asudeh and Dalrymple(2006)}]{asudeh_binding_2006}
A~Asudeh and M~Dalrymple. 2006.
\newblock Binding {Theory}.
\newblock In \emph{Encyploedia of Language \& Linguistics}, second edition. Elsevier Ltd.

\bibitem[{Carnie(2006)}]{carnie_island_2006}
A~Carnie. 2006.
\newblock Island {Constraints}.
\newblock In \emph{Encyploedia of Language \& Linguistics}, second edition. Elsevier Ltd.

\bibitem[{Castro et~al.(2009)Castro, Gómez, and Tejada}]{castro_polynomial_2009}
Javier Castro, Daniel Gómez, and Juan Tejada. 2009.
\newblock \href {https://doi.org/https://doi.org/10.1016/j.cor.2008.04.004} {Polynomial calculation of the {Shapley} value based on sampling}.
\newblock \emph{Computers \& Operations Research}, 36(5):1726--1730.

\bibitem[{Choenni et~al.(2023)Choenni, Shutova, and Garrette}]{choenni_examining_2023}
Rochelle Choenni, Ekaterina Shutova, and Dan Garrette. 2023.
\newblock \href {http://arxiv.org/abs/2311.08273} {Examining {Modularity} in {Multilingual} {LMs} via {Language}-{Specialized} {Subnetworks}}.
\newblock ArXiv:2311.08273 [cs].

\bibitem[{Clark et~al.(2019)Clark, Khandelwal, Levy, and Manning}]{clark-etal-2019-bert}
Kevin Clark, Urvashi Khandelwal, Omer Levy, and Christopher~D. Manning. 2019.
\newblock \href {https://doi.org/10.18653/v1/W19-4828} {What does {BERT} look at? an analysis of {BERT}{'}s attention}.
\newblock In \emph{Proceedings of the 2019 ACL Workshop BlackboxNLP: Analyzing and Interpreting Neural Networks for NLP}, pages 276--286, Florence, Italy. Association for Computational Linguistics.

\bibitem[{Dalvi et~al.(2019)Dalvi, Durrani, Sajjad, Belinkov, Bau, and Glass}]{dalvi2019one}
Fahim Dalvi, Nadir Durrani, Hassan Sajjad, Yonatan Belinkov, Anthony Bau, and James Glass. 2019.
\newblock What is one grain of sand in the desert? analyzing individual neurons in deep nlp models.
\newblock In \emph{Proceedings of the AAAI Conference on Artificial Intelligence}, volume~33, pages 6309--6317.

\bibitem[{Devlin et~al.(2019)Devlin, Chang, Lee, and Toutanova}]{devlin-etal-2019-bert}
Jacob Devlin, Ming-Wei Chang, Kenton Lee, and Kristina Toutanova. 2019.
\newblock \href {https://doi.org/10.18653/v1/N19-1423} {{BERT}: Pre-training of deep bidirectional transformers for language understanding}.
\newblock In \emph{Proceedings of the 2019 Conference of the North {A}merican Chapter of the Association for Computational Linguistics: Human Language Technologies, Volume 1 (Long and Short Papers)}, pages 4171--4186, Minneapolis, Minnesota. Association for Computational Linguistics.

\bibitem[{Dubinsky and Davies(2006)}]{dubinsky_control_2006}
S.W. Dubinsky and W.D. Davies. 2006.
\newblock \href {https://doi.org/10.1016/B0-08-044854-2/02000-9} {Control and {Raising}}.
\newblock In \emph{Encyclopedia of Language \& Linguistics}, second edition, pages 131--139. Elsevier Ltd.

\bibitem[{Dunn(1961)}]{dunn_multiple_1961}
Olive~Jean Dunn. 1961.
\newblock \href {https://doi.org/10.1080/01621459.1961.10482090} {Multiple {Comparisons} among {Means}}.
\newblock \emph{Journal of the American Statistical Association}, 56(293):52--64.

\bibitem[{Falk(2006)}]{falk_long-distance_2006}
Y~N Falk. 2006.
\newblock Long-{Distance} {Dependencies}.
\newblock In \emph{Encyclopedia of Language \& Linguistics}, second edition. Elsevier Ltd.

\bibitem[{Finlayson et~al.(2021)Finlayson, Mueller, Gehrmann, Shieber, Linzen, and Belinkov}]{finlayson-etal-2021-causal}
Matthew Finlayson, Aaron Mueller, Sebastian Gehrmann, Stuart Shieber, Tal Linzen, and Yonatan Belinkov. 2021.
\newblock \href {https://doi.org/10.18653/v1/2021.acl-long.144} {Causal analysis of syntactic agreement mechanisms in neural language models}.
\newblock In \emph{Proceedings of the 59th Annual Meeting of the Association for Computational Linguistics and the 11th International Joint Conference on Natural Language Processing (Volume 1: Long Papers)}, pages 1828--1843, Online. Association for Computational Linguistics.

\bibitem[{Foroutan et~al.(2022)Foroutan, Banaei, Lebret, Bosselut, and Aberer}]{foroutan-etal-2022-discovering}
Negar Foroutan, Mohammadreza Banaei, R{\'e}mi Lebret, Antoine Bosselut, and Karl Aberer. 2022.
\newblock \href {https://doi.org/10.18653/v1/2022.emnlp-main.513} {Discovering language-neutral sub-networks in multilingual language models}.
\newblock In \emph{Proceedings of the 2022 Conference on Empirical Methods in Natural Language Processing}, pages 7560--7575, Abu Dhabi, United Arab Emirates. Association for Computational Linguistics.

\bibitem[{Frankle and Carbin(2019)}]{frankle_lottery_2019}
Jonathan Frankle and Michael Carbin. 2019.
\newblock \href {https://doi.org/10.48550/arXiv.1803.03635} {The {Lottery} {Ticket} {Hypothesis}: {Finding} {Sparse}, {Trainable} {Neural} {Networks}}.
\newblock ArXiv:1803.03635 [cs].

\bibitem[{Ghorbani and Zou(2020)}]{ghorbani_neuron_2020}
Amirata Ghorbani and James~Y Zou. 2020.
\newblock \href {https://proceedings.neurips.cc/paper_files/paper/2020/hash/41c542dfe6e4fc3deb251d64cf6ed2e4-Abstract.html} {Neuron {Shapley}: {Discovering} the {Responsible} {Neurons}}.
\newblock In \emph{Advances in {Neural} {Information} {Processing} {Systems}}, volume~33, pages 5922--5932. Curran Associates, Inc.

\bibitem[{Held and Yang(2023)}]{held-yang-2023-shapley}
William Held and Diyi Yang. 2023.
\newblock \href {https://doi.org/10.18653/v1/2023.eacl-main.177} {Shapley head pruning: Identifying and removing interference in multilingual transformers}.
\newblock In \emph{Proceedings of the 17th Conference of the European Chapter of the Association for Computational Linguistics}, pages 2416--2427, Dubrovnik, Croatia. Association for Computational Linguistics.

\bibitem[{Hoeksema(2006)}]{hoeksema_polarity_2006}
J~Hoeksema. 2006.
\newblock Polarity {Items}.
\newblock In \emph{Encyclopedia of Language \& Linguistics}, second edition. Elsevier Ltd.

\bibitem[{Htut et~al.(2019)Htut, Phang, Bordia, and Bowman}]{htut_attention_2019}
Phu~Mon Htut, Jason Phang, Shikha Bordia, and Samuel~R. Bowman. 2019.
\newblock \href {https://doi.org/10.48550/arXiv.1911.12246} {Do {Attention} {Heads} in {BERT} {Track} {Syntactic} {Dependencies}?}
\newblock ArXiv:1911.12246.

\bibitem[{Hu et~al.(2021)Hu, Shen, Wallis, Allen-Zhu, Li, Wang, Wang, and Chen}]{hu_lora_2021}
Edward~J. Hu, Yelong Shen, Phillip Wallis, Zeyuan Allen-Zhu, Yuanzhi Li, Shean Wang, Lu~Wang, and Weizhu Chen. 2021.
\newblock \href {http://arxiv.org/abs/2106.09685} {{LoRA}: {Low}-{Rank} {Adaptation} of {Large} {Language} {Models}}.
\newblock ArXiv:2106.09685 [cs].

\bibitem[{Linzen and Baroni(2021)}]{linzen_syntactic_2021}
Tal Linzen and Marco Baroni. 2021.
\newblock \href {https://doi.org/10.1146/annurev-linguistics-032020-051035} {Syntactic {Structure} from {Deep} {Learning}}.
\newblock \emph{Annual Review of Linguistics}, 7(1):195--212.

\bibitem[{Liu et~al.(2019)Liu, Ott, Goyal, Du, Joshi, Chen, Levy, Lewis, Zettlemoyer, and Stoyanov}]{liu_roberta_2019}
Yinhan Liu, Myle Ott, Naman Goyal, Jingfei Du, Mandar Joshi, Danqi Chen, Omer Levy, Mike Lewis, Luke Zettlemoyer, and Veselin Stoyanov. 2019.
\newblock \href {http://arxiv.org/abs/1907.11692} {{RoBERTa}: {A} {Robustly} {Optimized} {BERT} {Pretraining} {Approach}}.
\newblock ArXiv:1907.11692 [cs].

\bibitem[{Liu and Chen(2023)}]{liu_picking_2023}
Zhengyuan Liu and Nancy~F. Chen. 2023.
\newblock \href {https://doi.org/10.1109/ICASSP49357.2023.10096717} {Picking the {Underused} {Heads}: {A} {Network} {Pruning} {Perspective} of {Attention} {Head} {Selection} for {Fusing} {Dialogue} {Coreference} {Information}}.
\newblock In \emph{{ICASSP} 2023 - 2023 {IEEE} {International} {Conference} on {Acoustics}, {Speech} and {Signal} {Processing} ({ICASSP})}, pages 1--5, Rhodes Island, Greece. IEEE.

\bibitem[{Maurer and Pontil(2009)}]{maurer_empirical_2009}
Andreas Maurer and Massimiliano Pontil. 2009.
\newblock \href {http://arxiv.org/abs/0907.3740} {Empirical {Bernstein} {Bounds} and {Sample} {Variance} {Penalization}}.
\newblock ArXiv:0907.3740 [stat].

\bibitem[{Mnih et~al.(2008)Mnih, Szepesvári, and Audibert}]{mnih_empirical_2008}
Volodymyr Mnih, Csaba Szepesvári, and Jean-Yves Audibert. 2008.
\newblock \href {https://doi.org/10.1145/1390156.1390241} {Empirical {Bernstein} stopping}.
\newblock In \emph{Proceedings of the 25th international conference on {Machine} learning - {ICML} '08}, pages 672--679, Helsinki, Finland. ACM Press.

\bibitem[{Mosca et~al.(2022)Mosca, Szigeti, Tragianni, Gallagher, and Groh}]{mosca-etal-2022-shap}
Edoardo Mosca, Ferenc Szigeti, Stella Tragianni, Daniel Gallagher, and Georg Groh. 2022.
\newblock \href {https://aclanthology.org/2022.coling-1.406} {{SHAP}-based explanation methods: A review for {NLP} interpretability}.
\newblock In \emph{Proceedings of the 29th International Conference on Computational Linguistics}, pages 4593--4603, Gyeongju, Republic of Korea. International Committee on Computational Linguistics.

\bibitem[{Mueller et~al.(2022)Mueller, Xia, and Linzen}]{mueller-etal-2022-causal}
Aaron Mueller, Yu~Xia, and Tal Linzen. 2022.
\newblock \href {https://doi.org/10.18653/v1/2022.conll-1.8} {Causal analysis of syntactic agreement neurons in multilingual language models}.
\newblock In \emph{Proceedings of the 26th Conference on Computational Natural Language Learning (CoNLL)}, pages 95--109, Abu Dhabi, United Arab Emirates (Hybrid). Association for Computational Linguistics.

\bibitem[{Pearl(2001)}]{pearl_direct_2001}
Judea Pearl. 2001.
\newblock Direct and indirect effects.
\newblock In \emph{Proceedings of the {Seventeenth} {Conference} on {Uncertainty} in {Artificial} {Intelligence}}, {UAI}'01, pages 411--420, San Francisco, CA, USA. Morgan Kaufmann Publishers Inc.
\newblock Event-place: Seattle, Washington.

\bibitem[{Pfeiffer et~al.(2024)Pfeiffer, Ruder, Vulić, and Ponti}]{pfeiffer_modular_2024}
Jonas Pfeiffer, Sebastian Ruder, Ivan Vulić, and Edoardo~Maria Ponti. 2024.
\newblock \href {http://arxiv.org/abs/2302.11529} {Modular {Deep} {Learning}}.
\newblock ArXiv:2302.11529 [cs].

\bibitem[{Shapley(1953)}]{shapley_17_1953}
L.~S. Shapley. 1953.
\newblock \href {https://doi.org/doi:10.1515/9781400881970-018} {17. {A} {Value} for n-{Person} {Games}}.
\newblock In Harold~William Kuhn and Albert~William Tucker, editors, \emph{Contributions to the {Theory} of {Games} ({AM}-28), {Volume} {II}}, pages 307--318. Princeton University Press, Princeton.

\bibitem[{Someya and Oseki(2023)}]{someya-oseki-2023-jblimp}
Taiga Someya and Yohei Oseki. 2023.
\newblock \href {https://doi.org/10.18653/v1/2023.findings-eacl.117} {{JBL}i{MP}: {J}apanese benchmark of linguistic minimal pairs}.
\newblock In \emph{Findings of the Association for Computational Linguistics: EACL 2023}, pages 1581--1594, Dubrovnik, Croatia. Association for Computational Linguistics.

\bibitem[{Song et~al.(2022)Song, Krishna, Bhatt, and Iyyer}]{song-etal-2022-sling}
Yixiao Song, Kalpesh Krishna, Rajesh Bhatt, and Mohit Iyyer. 2022.
\newblock \href {https://doi.org/10.18653/v1/2022.emnlp-main.305} {{SLING}: {S}ino linguistic evaluation of large language models}.
\newblock In \emph{Proceedings of the 2022 Conference on Empirical Methods in Natural Language Processing}, pages 4606--4634, Abu Dhabi, United Arab Emirates. Association for Computational Linguistics.

\bibitem[{Stanczak et~al.(2022)Stanczak, Ponti, Torroba~Hennigen, Cotterell, and Augenstein}]{stanczak-etal-2022-neurons}
Karolina Stanczak, Edoardo Ponti, Lucas Torroba~Hennigen, Ryan Cotterell, and Isabelle Augenstein. 2022.
\newblock \href {https://doi.org/10.18653/v1/2022.naacl-main.114} {Same neurons, different languages: Probing morphosyntax in multilingual pre-trained models}.
\newblock In \emph{Proceedings of the 2022 Conference of the North American Chapter of the Association for Computational Linguistics: Human Language Technologies}, pages 1589--1598, Seattle, United States. Association for Computational Linguistics.

\bibitem[{Taktasheva et~al.(2024)Taktasheva, Bazhukov, Koncha, Fenogenova, Artemova, and Mikhailov}]{taktasheva_rublimp_2024}
Ekaterina Taktasheva, Maxim Bazhukov, Kirill Koncha, Alena Fenogenova, Ekaterina Artemova, and Vladislav Mikhailov. 2024.
\newblock \href {http://arxiv.org/abs/2406.19232} {{RuBLiMP}: {Russian} {Benchmark} of {Linguistic} {Minimal} {Pairs}}.
\newblock ArXiv:2406.19232 [cs].

\bibitem[{Torroba~Hennigen et~al.(2020)Torroba~Hennigen, Williams, and Cotterell}]{torroba-hennigen-etal-2020-intrinsic}
Lucas Torroba~Hennigen, Adina Williams, and Ryan Cotterell. 2020.
\newblock \href {https://doi.org/10.18653/v1/2020.emnlp-main.15} {Intrinsic probing through dimension selection}.
\newblock In \emph{Proceedings of the 2020 Conference on Empirical Methods in Natural Language Processing (EMNLP)}, pages 197--216, Online. Association for Computational Linguistics.

\bibitem[{Vig et~al.(2020)Vig, Gehrmann, Belinkov, Qian, Nevo, Singer, and Shieber}]{vig_investigating_2020}
Jesse Vig, Sebastian Gehrmann, Yonatan Belinkov, Sharon Qian, Daniel Nevo, Yaron Singer, and Stuart Shieber. 2020.
\newblock \href {https://proceedings.neurips.cc/paper_files/paper/2020/hash/92650b2e92217715fe312e6fa7b90d82-Abstract.html} {Investigating {Gender} {Bias} in {Language} {Models} {Using} {Causal} {Mediation} {Analysis}}.
\newblock In \emph{Advances in {Neural} {Information} {Processing} {Systems}}, volume~33, pages 12388--12401. Curran Associates, Inc.

\bibitem[{Warstadt et~al.(2020)Warstadt, Parrish, Liu, Mohananey, Peng, Wang, and Bowman}]{warstadt-etal-2020-blimp}
Alex Warstadt, Alicia Parrish, Haokun Liu, Anhad Mohananey, Wei Peng, Sheng-Fu Wang, and Samuel~R. Bowman. 2020.
\newblock \href {https://aclanthology.org/2020.scil-1.47} {{BL}i{MP}: A benchmark of linguistic minimal pairs for {E}nglish}.
\newblock In \emph{Proceedings of the Society for Computation in Linguistics 2020}, pages 409--410, New York, New York. Association for Computational Linguistics.

\bibitem[{Wilcox et~al.(2024)Wilcox, Futrell, and Levy}]{wilcox_using_2024}
Ethan~Gotlieb Wilcox, Richard Futrell, and Roger Levy. 2024.
\newblock \href {https://doi.org/10.1162/ling_a_00491} {Using {Computational} {Models} to {Test} {Syntactic} {Learnability}}.
\newblock \emph{Linguistic Inquiry}, 55(4):805--848.

\bibitem[{Winkler(2006)}]{winkler_ellipsis_2006}
S~Winkler. 2006.
\newblock Ellipsis.
\newblock In \emph{Encyclopedia of Language \& Linguistics}, second edition. Elsevier Ltd.

\end{thebibliography}
\bibliographystyle{acl_natbib}

\newpage

\appendix

\section{All BLiMP Paradigms}
\label{sec:appendix_blimp}

See Table \ref{tab:all_blimp_paradigms} for a list of all minimal pair paradigms and examples in the BLiMP dataset \citep{warstadt-etal-2020-blimp}, organised into their relevant linguistic categories.

\begin{table*}[]
    \centering
    \resizebox{\textwidth}{!}{
    \begin{tabular}{ll}
    \hline
    BLiMP Paradigm & Grammatical/Ungrammatical Examples \\
    \hline
    \multicolumn{2}{c}{\textsc{Anaphor agr.}} \\
    \hline
    Anaphor gender agreement & A girl couldn't reveal herself/*himself. \\
    Anaphor number agreement & Thomas complained about himself/*themselves. \\
    \hline
    \multicolumn{2}{c}{\textsc{Arg. structure}} \\
    \hline
    Causative & Aaron breaks/*appeared the glass. \\
    Drop argument & Travis is touring/*Travis is revealing. \\
    Inchoative & Patricia had changed./*Patricia had forgotten. \\
    Intransitive & The screen does brighten/*resemble. \\
    Passive 1 & Tracy isn't fired/*muttered by Jodi's daughter. \\
    Passive 2 & Steve isn't disliked/*lied. \\
    Transitive & Diane watched/*screamed Alan. \\
    \hline
    \multicolumn{2}{c}{\textsc{Binding}} \\
    \hline
    Principle A (c-command) & A girl that wouldn't watch Omar questions herself/*himself. \\
    \qquad " \qquad (case 1) & The teenagers explain that they/*themselves aren't breaking all glasses. \\
    \qquad " \qquad (case 2) & Eric imagines himself taking every rug/*Eric imagines himself took every rug. \\
    \qquad " \qquad (domain 1) & Carla had explained that Samuel has discussed her/*herself. \\
    \qquad " \qquad (domain 2) & James says Kayla helped herself/*himself. \\
    \qquad " \qquad (domain 3) & Gina explains Alan fires himself/*Alan explains Gina fires himself. \\
    \qquad " \qquad (reconstruction) & It's himself that this cashier attacked/*It's himself that attacked this cashier. \\
    \hline
    \multicolumn{2}{c}{\textsc{Control/raising}} \\
    \hline
    Existential `there' (object raising) & Frank judged /*compelled there to be a photograph of Michael looking like Sherry. \\
    \qquad " \qquad\qquad (subject raising) & There was bound to be a fish escaping/*There was unable to be a fish escaping. \\
    Expletive `it' object raising & This cashier had ascertained/*can't press it to be not so interesting that Anna painted. \\
    `Tough' vs raising (1) & Julia wasn't fun/*unlikely to talk to. \\
    \qquad " \qquad\qquad (2) & Bruce was sure/*annoying to remember Gerald. \\
    \hline
    \multicolumn{2}{c}{\textsc{Det.-noun agr.}} \\
    \hline
    Determiner-noun agreement (1) & Raymond is selling this sketch/*sketches. \\
    \qquad " \qquad\qquad\qquad\qquad (2) & Tracy passed these/*this art galleries. \\
    \qquad " \qquad\qquad\qquad\qquad (irregular 1) & The driver reveals these/*this mice. \\
    \qquad " \qquad\qquad\qquad\qquad (irregular 2) & Natalie describes this/*these child. \\
    \qquad " \qquad\qquad\qquad\qquad (with adjective 1) & Many men have these messy cups/*cup. \\
    \qquad " \qquad\qquad\qquad\qquad (with adjective 2) & Donna might hire this/*these serious actress. \\
    \qquad " \qquad\qquad\qquad\qquad (with adjective, irregular 1) & Heidi returns to that big woman/*women. \\
    \qquad " \qquad\qquad\qquad\qquad (with adjective, irregular 2) & Denise did confuse that/*those important women. \\
    \hline
    \multicolumn{2}{c}{\textsc{Ellipsis}} \\
    \hline
    Ellipsis N-bar (1) & This print scares a lot of busy senators and Benjamin scares a few/*This print scares a lot of senators and Benjamin scares a few busy. \\
    \qquad " \qquad\qquad (2) & Vincent wore one shirt and Matt wore some big shirt/*Vincent wore one hidden shirt and Matt wore some big. \\
    \hline
    \multicolumn{2}{c}{\textsc{Filler-gap dep.}} \\
    \hline
    Wh-questions (object gap) & Joel discovered the vase that Patricia might take/*Joel discovered what Patricia might take the vase. \\
    \qquad " \qquad (subject gap) & Leslie remembered some guest that has bothered women./*Leslie remembered who some guest has bothered women. \\
    \qquad " \qquad (subject gap, long distance) & Regina sees that candle that Steve lifts that might impress every doctor./*Regina sees who that candle that Steve lifts might impress every doctor. \\
    Wh vs `that' (no gap) & Mark figured out that/*who most governments appreciate Steven. \\
    \qquad " \qquad (no gap, long distance) & Eva discovered that/*who all pedestrians that have performed upset Candice. \\
    \qquad " \qquad (with gap) & Wayne has revealed who/*that most hospitals admired. \\
    \qquad " \qquad (with gap, long distance) & Kenneth investigated who/*that the cashiers that perform cared for. \\
    \hline
    \multicolumn{2}{c}{\textsc{Irregular forms}} \\
    \hline
    Irregular past participle (adjectives) & The broken/*broke mirrors were blue. \\
    \qquad " \qquad (verbs) & The Borgias wore/*worn a lot of scarves. \\
    \hline
    \multicolumn{2}{c}{\textsc{Island effects}} \\
    \hline
    Adjunct island & Who should Derek hug after shocking Richard?/*Who should Derek hug Richard after shocking? \\
    Complex NP island & What can't a guest who would like some actor argue about?/*What can't some actor argue about a guest who would like? \\
    Coordinate structure constraint (complex left branch) & Which teenagers had Tamara hired and Grace fired?/*Which hard Tamara hired teenagers and Grace fired? \\
    \qquad " \qquad\qquad\qquad\qquad\qquad (object extraction) & What had Russel and Douglas attacked?/*What had Russel attacked and Douglas? \\
    Left branch island (echo question) & Irene had messed up whose rug?/*Whose had Irene messed up rug? \\
    \qquad " \qquad\qquad (simple question) & Whose museums had Dana alarmed?/*Whose had Dana alarmed museums? \\
    Sentential subject island & Who should pedestrians' curing Deanna scare/*Who should pedestrians' curing scare Deanna. \\
    Wh-island & Who isn't Craig realising he/*who kisses? \\
    \hline
    \multicolumn{2}{c}{\textsc{NPI licensing}} \\
    \hline
    NPI licensor present (matrix question) & Had Bruce ever played? / *Bruce had ever played. \\
    \qquad " \qquad\qquad (`only') & Only/*Even Bill would ever complain. \\
    \qquad " \qquad\qquad (sentential negation) & Teresa had not/*probably every sold a movie theater. \\
    NPI present (1) & Even Suzanne has really/*ever joked around. \\
    \qquad " \qquad (2) & Tamara really/*ever exited these mountains. \\
    NPI scope (`only') & Only many people who George likes ever clashed./*Many people who only George likes ever clashed. \\
    \qquad " \qquad (sentential negation) & Every coat that did scare Nina has not ever wrinkled/*Every coat that did not scare Nina has ever wrinkled. \\
    \hline
    \multicolumn{2}{c}{\textsc{Quantifiers}} \\
    \hline
    Existential `there' quantifiers (1) & There was a/*each documentary about music irritating Allison. \\
    \qquad " \qquad\qquad\qquad\qquad (2) & All dancers are there talking to Pamela./*There are all dancers talking to Pamela. \\
    Superlative quantifiers (1) & No girl attacked fewer/*at most than two waiters. \\
    \qquad " \qquad\qquad\qquad (2) & The/*No hospital had fired at most four people. \\
    \hline
    \multicolumn{2}{c}{\textsc{S-selection}} \\
    \hline
    Animate subject (passive) & Lisa was kissed by the boys/*blouses. \\
    \qquad " \qquad\qquad (transitive) & Phillip/*This pasta can talk to those waitresses. \\
    \hline
    \multicolumn{2}{c}{\textsc{Subject-verb agr.}} \\
    \hline
    Distractor agreement (relational noun) & A story about the Balkans doesn't/*don't irritate a person. \\
    \qquad " \qquad\qquad\qquad (relative clause) & Boys that aren't disturbing Natalie suffer/*suffers. \\
    Irregular plural subject verb agreement (1) & This goose isn't/*weren't bothering Edward. \\
    \qquad " \qquad\qquad\qquad\qquad\qquad\qquad (2) & The people/*person conspire. \\
    Regular plural subject verb agreement (1) & The cups alarm/*alarms Angela. \\
    \qquad " \qquad\qquad\qquad\qquad\qquad\qquad (2) & The students/*student perform. \\
    \bottomrule
    \end{tabular}
    }
    \caption{A list of all BLiMP minimal pair paradigms and examples, organised according to their respective linguistic categories.}
    \label{tab:all_blimp_paradigms}
\end{table*}

\section{Cluster Purity}
\label{sec:appendix_purity}

\begin{table}[H]
    \centering
    \begin{tabular}{lr}
    \toprule
    Clusters & Purity: $\mu$ ($\sigma$) \\
    \bottomrule
    \textsc{BERT Reference} & \\
    \qquad \underline{BERT clusters} & \underline{0.497 (0.045)} \\
    \qquad RoBERTa clusters & 0.530 (0.046) \\
    \qquad Random clusters & 0.277 (0.027) \\
    \cmidrule(rl){1-2}
    \textsc{RoBERTa Reference} & \\
    \qquad BERT clusters & 0.532 (0.044) \\
    \qquad \underline{RoBERTa clusters} & \underline{0.765 (0.063)} \\
    \qquad Random clusters & 0.285 (0.028) \\
    \bottomrule
    \end{tabular}
    \caption{
    Mean ($\mu$) purity scores and standard deviations ($\sigma$) across $k$-means runs measured against our BERT and RoBERTa reference clusters.
    Each purity score represents comparison with 100 individual runs, within-model comparisons are underlined.
    }
    \label{tab:purity}
\end{table}

Since $k$-means clustering is a stochastic process, SHV clusters may differ between runs of the algorithm.
We focus on showing how qualitative and quantitative analysis can reveal shared patterns between clusters in general, thus ensuring clustering consistency is not our goal.
Nevertheless, we evaluate do carry out intrinsic evaluation of the method using \textit{purity} as a metric.
Given $N$ data points, a set of clusters $\Omega = \{\omega_1, \omega_2, \ldots, \omega_K\}$, and a set of classes $\mathbb{C} = \{c_1, c_2, \ldots, c_J\}$, purity is calculated through assigning each cluster to the class most frequent in the cluster, and then measuring the number of correctly assigned data points divided by $N$, see Equation \ref{eq:purity}.

\begin{equation}
    \label{eq:purity}
    \operatorname{purity}(\Omega, \mathbb{C}) = \frac{1}{N} \sum_k \operatorname{max}_j | \omega_k \cap c_j |
\end{equation}

Purity falls between 0 (no match between clusters) and 1 (perfect match), and it is typically calculated against a gold standard cluster set.
As we do not have gold clusters, we measure the purity using reference clusters for both BERT and RoBERTa (see the clusters in Section \ref{sec:results}).
Since cluster labels may also change between runs, we cannot simply use cluster IDs as gold labels even using the reference clusters.
Instead, we aim to align clusters using the Hungarian (or Kuhn-Munkres) algorithm that matches clusters by maximising shared datapoints between them.

Table \ref{tab:purity} shows purity scores between BERT and RoBERTa clusters and three sets of 100 different cluster sets: these include other runs of $k$-means clustering on the BERT and RoBERTa SHVs, and randomly assigned clusters.
It is clear that the cohesion of clusters as measured in terms of purity across and within models far exceeds the cohesion between the reference clusters and the random clusters.

\section{All Clusters}
\label{sec:appendix_clusters}

See Tables \ref{tab:bert_roberta_clusters_1} and \ref{tab:bert_roberta_clusters_2} that list all BERT and RoBERTa clusters, representing both matching and (underlined in the tables) and not matching ones.

\begin{table*}[]
    \centering
    \resizebox{0.7\textwidth}{!}{
    \begin{tabular}{lllc}
    \toprule
     & BLiMP Paradigm  & Linguistics Term & M \\
    \midrule
    \multirow{8}{*}{
    \rotatebox{90}{\textbf{\underline{NPI Cluster}}}
    } & NPI present (1) & \multirow{7}{*}{\textsc{NPI licensing}} & $\lozenge$ \\
    & NPI present (2) & & \textcolor{my_turqoise}{$\blacklozenge$} \\
    & NPI scope (`only') & & \textcolor{my_purple}{$\blacklozenge$} \\
    & NPI scope (sentential negation) & & $\lozenge$ \\
    & NPI licensor present (`only') & & $\lozenge$ \\
    & \qquad " \qquad (matrix question) & & $\lozenge$ \\
    & \qquad " \qquad (sentential negation) & & $\lozenge$ \\
    \cmidrule(rl){2-4}
    & Irregular past participle verbs & \textsc{Irregular forms} & \textcolor{my_turqoise}{$\blacklozenge$} \\
    \midrule
    \multirow{7}{*}{
    \rotatebox{90}{\textbf{\underline{Island Effects}}}
    } & Adjunct island & \multirow{7}{*}{\textsc{Island effects}} & $\lozenge$ \\
    & Complex NP island & & $\lozenge$ \\
    & Coordinate structure constraint (complex left branch) & & $\lozenge$ \\
    & \qquad " \qquad (object extraction) & & \textcolor{my_purple}{$\blacklozenge$} \\
    & Left branch island (simple question) & & $\lozenge$ \\
    & Left branch island (echo question) & & $\lozenge$ \\
    & Wh-island & & \textcolor{my_purple}{$\blacklozenge$} \\
    \midrule
    \multirow{2}{*}{
    \rotatebox{90}{\makecell{\textbf{\underline{Quan-}} \\ \textbf{\underline{tifiers}}}}
    } & Superlative quantifiers 1 & \multirow{2}{*}{\textsc{Quantifiers}} & $\lozenge$ \\
    & Superlative quantifiers 2 & & $\lozenge$ \\
    \midrule
    \multirow{7}{*}{
    \rotatebox{90}{\textbf{\underline{Binding*}}}
    } & Anaphor gender agreement & \textsc{Anaphor agr.} & \textcolor{my_purple}{$\blacklozenge$} \\
    \cmidrule(rl){2-4}
    & Animate subject trans. & \textsc{S-selection} & \textcolor{my_turqoise}{$\blacklozenge$} \\
    \cmidrule(rl){2-4}
    & Principle A (case 1) & \multirow{5}{*}{\textsc{Binding}} & \textcolor{my_turqoise}{$\blacklozenge$} \\
    & \qquad " \qquad (domain 1) & & $\lozenge$ \\
    & \qquad " \qquad \underline{(domain 2)} & & $\lozenge$ \\
    & \qquad " \qquad \underline{(domain 3)} & & $\lozenge$ \\
    & \qquad " \qquad (c-command) & & $\lozenge$ \\
    \midrule
    \multirow{11}{*}{
    \rotatebox{90}{\textbf{\underline{Filler-Gap}}}
    } & Ellipsis N-bar (2) & \textsc{Ellipsis} & \textcolor{my_turqoise}{$\blacklozenge$} \\
    \cmidrule(rl){2-4}
    & Existential `there' (subject raising) & \multirow{2}{*}{\textsc{Control/raising}} & \textcolor{my_turqoise}{$\blacklozenge$} \\
    & Tough vs raising (2) & & \textcolor{my_purple}{$\blacklozenge$} \\
    \cmidrule(rl){2-4}
    & Principle A (case 1) & \multirow{3}{*}{\textsc{Binding}} & \textcolor{my_purple}{$\blacklozenge$} \\
    & \qquad " \qquad (case 2) & & $\lozenge$ \\
    & \qquad " \qquad (reconstruction) & & $\lozenge$ \\
    \cmidrule(rl){2-4}
    & Wh-questions (object gap) & \multirow{5}{*}{\textsc{Filler-gap dep.}} & $\lozenge$ \\
    & \qquad " \qquad (subject gap) & & $\lozenge$ \\
    & \qquad " \qquad (subject gap, long distance) & & $\lozenge$ \\
    & Wh vs `that' (no gap) & & $\lozenge$ \\
    & \qquad " \qquad (no gap, long distance) & & $\lozenge$ \\
    \midrule
    \multirow{5}{*}{
    \rotatebox{90}{\textbf{\underline{Wh vs `That'}}}
    } & Inchoative & \multirow{2}{*}{\textsc{Arg. structure}} & \textcolor{my_purple}{$\blacklozenge$} \\
    & Intransitive & & \textcolor{my_purple}{$\blacklozenge$} \\
    \cmidrule(rl){2-4}
    & Tough vs raising (1) & \textsc{Control/raising} & \textcolor{my_purple}{$\blacklozenge$} \\
    \cmidrule(rl){2-4}
    & Wh vs `that' (with gap) & \multirow{2}{*}{\textsc{Filler-gap dep.}} & $\lozenge$ \\
    & \qquad " \qquad (with gap, long distance) & & $\lozenge$ \\
    \midrule
    \multirow{8}{*}{
    \rotatebox{90}{\textbf{BERT 1}}
    } & Anaphor gender agreement & \textsc{Anaphor agr.} & \textcolor{my_turqoise}{$\blacklozenge$} \\
    \cmidrule(rl){2-4}
    & Distractor agreement (relational noun) & \multirow{3}{*}{\textsc{Subject-verb agr.}} & \textcolor{my_turqoise}{$\blacklozenge$} \\
    & \qquad " \qquad (relative clause) & & \textcolor{my_turqoise}{$\blacklozenge$} \\
    & Irregular plural subject verb agreement (1) & & \textcolor{my_turqoise}{$\blacklozenge$} \\
    \cmidrule(rl){2-4}
    & Existential `there' (object raising) & \multirow{2}{*}{\textsc{Control/raising}} & \textcolor{my_turqoise}{$\blacklozenge$} \\
    & Expletive `it' (object raising) & & \textcolor{my_turqoise}{$\blacklozenge$} \\
    \cmidrule(rl){2-4}
    & NPI scope (`only') & \textsc{NPI licensing} & \textcolor{my_turqoise}{$\blacklozenge$} \\
    \cmidrule(rl){2-4}
    & Sentential subject island & \textsc{Island effects} & \textcolor{my_turqoise}{$\blacklozenge$} \\
    \midrule
    \multirow{3}{*}{
    \rotatebox{90}{\textbf{BERT 2}}
    } & Irregular past participle adjectives & \textsc{Irregular forms} & \textcolor{my_turqoise}{$\blacklozenge$} \\
    \cmidrule(rl){2-4}
    & Passive (1) & \multirow{2}{*}{\textsc{Arg. structure}} & \textcolor{my_turqoise}{$\blacklozenge$} \\
    & Passive (2) & & \textcolor{my_turqoise}{$\blacklozenge$} \\
    \midrule
    \multirow{3}{*}{
    \rotatebox{90}{\textbf{\scriptsize{RoBERTa 1}}}
    } & Animate subject (passive) & \multirow{2}{*}{\textsc{S-selection}} & \textcolor{my_purple}{$\blacklozenge$} \\
    & \qquad " \qquad\qquad (transitive) & & \textcolor{my_purple}{$\blacklozenge$} \\
    \cmidrule(rl){2-4}
    & Passive (1) & \textsc{Arg. structure} & \textcolor{my_purple}{$\blacklozenge$} \\
    \bottomrule
    \end{tabular}
    }
    \caption{
    BLiMP paradigms in the relevant BERT (\textcolor{my_turqoise}{$\blacklozenge$}) and RoBERTa (\textcolor{my_purple}{$\blacklozenge$}) clusters.
    Paradigms that appear in the clusters of both models are marked with $\lozenge$.
    Clusters without obvious organising principles are marked with model names and numbers instead of labels.
    }
    \label{tab:bert_roberta_clusters_1}
\end{table*}

\begin{table*}[]
    \centering
    \resizebox{0.65\textwidth}{!}{
    \begin{tabular}{lllc}
    \toprule
     & BLiMP Paradigm  & Linguistics Term & M \\
    \midrule
    \multirow{24}{*}{
    \rotatebox{90}{\textbf{BERT 3}}
    } & Anaphor number agreement & \textsc{Anaphor agr.} & \textcolor{my_turqoise}{$\blacklozenge$} \\
    \cmidrule(rl){2-4}
    & Animate subject (passive) & \textsc{S-selection} & \textcolor{my_turqoise}{$\blacklozenge$} \\
    \cmidrule(rl){2-4}
    & Causative & \multirow{5}{*}{\textsc{Arg. structure}} & \textcolor{my_turqoise}{$\blacklozenge$} \\
    & Drop argument & & \textcolor{my_turqoise}{$\blacklozenge$} \\
    & Inchoative & & \textcolor{my_turqoise}{$\blacklozenge$} \\
    & Intransitive & & \textcolor{my_turqoise}{$\blacklozenge$} \\
    & Transitive & & \textcolor{my_turqoise}{$\blacklozenge$} \\
    \cmidrule(rl){2-4}
    & Coordinate structure constraint (object extraction) & \multirow{2}{*}{\textsc{Island effects}} & \textcolor{my_turqoise}{$\blacklozenge$} \\
    & Wh-island & & \textcolor{my_turqoise}{$\blacklozenge$} \\
    \cmidrule(rl){2-4}
     & Determiner-noun agr. (1) & \multirow{8}{*}{\textsc{Det.-noun agr.}} & \textcolor{my_turqoise}{$\blacklozenge$} \\
    & \qquad " \qquad (2) & & \textcolor{my_turqoise}{$\blacklozenge$} \\
    & \qquad " \qquad (irregular, 1) & & \textcolor{my_turqoise}{$\blacklozenge$} \\
    & \qquad " \qquad (irregular, 2) & & \textcolor{my_turqoise}{$\blacklozenge$} \\
    & \qquad " \qquad (with adjective, 1) & & \textcolor{my_turqoise}{$\blacklozenge$} \\
    & \qquad " \qquad (with adjective, 2) & & \textcolor{my_turqoise}{$\blacklozenge$} \\
    & \qquad " \qquad  (with adjective, irregular 1) & & \textcolor{my_turqoise}{$\blacklozenge$} \\
    & \qquad " \qquad (with adjective, irregular 2) & & \textcolor{my_turqoise}{$\blacklozenge$} \\
    \cmidrule(rl){2-4}
    & Ellipsis N-bar (1) & \textsc{Ellipsis} & \textcolor{my_turqoise}{$\blacklozenge$} \\
    \cmidrule(rl){2-4}
    & Existential `there' quantifiers (1) & \multirow{2}{*}{\textsc{Quantifiers}} & \textcolor{my_turqoise}{$\blacklozenge$} \\
    & \qquad " \qquad\qquad\qquad\qquad\quad (2) & & \textcolor{my_turqoise}{$\blacklozenge$} \\
    \cmidrule(rl){2-4}
    & Irregular plural subject-verb agreement (2) & \multirow{3}{*}{\textsc{Subject-verb agr.}} & \textcolor{my_turqoise}{$\blacklozenge$} \\
    & Regular plural subject-verb agreement (1) & & \textcolor{my_turqoise}{$\blacklozenge$} \\
    & \qquad " \qquad\qquad\qquad\qquad\qquad\qquad (2) & & \textcolor{my_turqoise}{$\blacklozenge$} \\
    \cmidrule(rl){2-4}
    & `Tough' vs raising (1) & \multirow{2}{*}{\textsc{Control/raising}} & \textcolor{my_turqoise}{$\blacklozenge$} \\
    & \qquad " \qquad\qquad (2) & & \textcolor{my_turqoise}{$\blacklozenge$} \\
    \midrule
    \multirow{12}{*}{
    \rotatebox{90}{\textbf{Det.-Noun Cluster (RoBERTa)}}
    } & Determiner-noun agr. (1) & \multirow{8}{*}{\textsc{Det.-noun agr.}} & \textcolor{my_purple}{$\blacklozenge$} \\
    & \qquad " \qquad (2) & & \textcolor{my_purple}{$\blacklozenge$} \\
    & \qquad " \qquad (irregular, 1) & & \textcolor{my_purple}{$\blacklozenge$} \\
    & \qquad " \qquad (irregular, 2) & & \textcolor{my_purple}{$\blacklozenge$} \\
    & \qquad " \qquad (with adjective, 1) & & \textcolor{my_purple}{$\blacklozenge$} \\
    & \qquad " \qquad (with adjective, 2) & & \textcolor{my_purple}{$\blacklozenge$} \\
    & \qquad " \qquad  (with adjective, irregular 1) & & \textcolor{my_purple}{$\blacklozenge$} \\
    & \qquad " \qquad (with adjective, irregular 2) & & \textcolor{my_purple}{$\blacklozenge$} \\
    \cmidrule(rl){2-4}
    & Distractor agreement (relational noun) & \multirow{3}{*}{\textsc{Subject-verb agr.}} & \textcolor{my_purple}{$\blacklozenge$} \\
    & Regular plural subject-verb agr. (1) & & \textcolor{my_purple}{$\blacklozenge$} \\
    & Regular plural subject-verb agr. (2) & & \textcolor{my_purple}{$\blacklozenge$} \\
    \cmidrule(rl){2-4}
    & Transitive & \textsc{Arg. structure} & \textcolor{my_purple}{$\blacklozenge$} \\
    \midrule
    \multirow{12}{*}{
    \rotatebox{90}{\textbf{RoBERTa 2}}
    } & Anaphor number agreement & \textsc{Anaphor agr.} & \textcolor{my_purple}{$\blacklozenge$} \\
    \cmidrule(rl){2-4}
    & Causative & \multirow{3}{*}{\textsc{Arg. structure}} & \textcolor{my_purple}{$\blacklozenge$} \\
    & Drop argument & & \textcolor{my_purple}{$\blacklozenge$} \\
    & Passive (2) & & \textcolor{my_purple}{$\blacklozenge$} \\
    \cmidrule(rl){2-4}
    & Ellipsis N-bar (1) & \multirow{2}{*}{\textsc{Ellipsis}} & \textcolor{my_purple}{$\blacklozenge$} \\
    & \qquad " \qquad\quad (2) & & \textcolor{my_purple}{$\blacklozenge$} \\
    \cmidrule(rl){2-4}
    & Irregular past participle (adjectives) & \multirow{2}{*}{\textsc{Irregular forms}} & \textcolor{my_purple}{$\blacklozenge$} \\
    & \qquad " \qquad\qquad\qquad (verbs) & & \textcolor{my_purple}{$\blacklozenge$} \\
    \cmidrule(rl){2-4}
    & Irregular plural subject-verb agreement (1) & \multirow{2}{*}{\textsc{Subject-verb agr.}} & \textcolor{my_purple}{$\blacklozenge$} \\
    & \qquad " \qquad\qquad\qquad\qquad\qquad\qquad\quad (2) & & \textcolor{my_purple}{$\blacklozenge$} \\
    \cmidrule(rl){2-4}
    & NPI present (2) & \textsc{NPI licensing} & \textcolor{my_purple}{$\blacklozenge$} \\
    \cmidrule(rl){2-4}
    & Sentential subject island & \textsc{Island effects} & \textcolor{my_purple}{$\blacklozenge$} \\
    \midrule
    \multirow{6}{*}{
    \rotatebox{90}{\textbf{RoBERTa 3}}
    } & Distractor agreement (relative clause) & \textsc{Subject-verb agr.} & \textcolor{my_purple}{$\blacklozenge$} \\
    \cmidrule(rl){2-4}
    & Existential `there' object raising & \textsc{Control/raising} & \textcolor{my_purple}{$\blacklozenge$} \\
    & Existential `there' quantifiers (1) & \multirow{2}{*}{\textsc{Quantifiers}} & \textcolor{my_purple}{$\blacklozenge$} \\
    & \qquad " \qquad (2) & & \textcolor{my_purple}{$\blacklozenge$} \\
    \cmidrule(rl){2-4}
    & Existential `there' subject raising & \multirow{2}{*}{\textsc{Control/raising}} & \textcolor{my_purple}{$\blacklozenge$} \\
    & Expletive `it' object raising & & \textcolor{my_purple}{$\blacklozenge$} \\
    \bottomrule
    \end{tabular}
    }
    \caption{
    BLiMP paradigms in the relevant BERT (\textcolor{my_turqoise}{$\blacklozenge$}) and RoBERTa (\textcolor{my_purple}{$\blacklozenge$}) clusters (continued).
    Paradigms that appear in the clusters of both models are marked with $\lozenge$.
    Clusters without obvious organising principles are marked with model names and numbers instead of labels.
    }
    \label{tab:bert_roberta_clusters_2}
\end{table*}

\section{T-Test on Pruning In- and Out-of-Cluster}
\label{sec:appendix_ttest}

\begin{table}[H]
    \centering
    \begin{tabular}{lrr}
    \toprule
    Cluster & T-stat. & P-value \\
    \toprule
    \textbf{NPI} & $-4.809$ & $1.12\textrm{e}{-05}$ \\
    Island & $0.995$ & $0.329$ \\
    Quantifiers & $-2.053$ & $0.113$ \\
    \textbf{Binding} & $-8.719$ & $7.97\textrm{e}{-11}$ \\
    \textbf{Filler-gap} & $-4.885$ & $2.94\textrm{e}{-06}$ \\
    Wh vs `that' & $-1.855$ & $0.157$ \\
    \bottomrule
    \end{tabular}
    \caption{Results of the T-test between in-cluster and out-of-cluster pruning with the BERT model.}
    \label{tab:ttest}
\end{table}

Table \ref{tab:ttest} shows T-test statistics and P-values -- adjusted using the Bonferroni correction \citep{dunn_multiple_1961} -- between accuracy changes in the in-cluster and out-of-cluster pruning of relevant clusters with the BERT model.
The \textbf{NPI}, \textbf{Binding}, and \textbf{Filler-gap} clusters pass the T-test, i.e., the differences between the distribution of in- and out-of-cluster values are statistically significant.
This indicates a more drastic consequence of pruning within BLiMP paradigms in these clusters than when using prune masks from paradigms outside of the clusters.

This is not the case in the other three clusters.
However, both the \textbf{Quantifiers} and \textbf{Wh vs `that'} clusters contain only 2-2 paradigms, meaning that in-cluster pruning involves only four datapoints.
This makes the success of any statistical analysis questionable.
Finally, the \textbf{Island} cluster contains only 5 out of 8 \textsc{Island effects} paradigms.
Additionally, it is also possible that the attention heads responsible for processing these paradigms are not so well localisable as with other paradigms.
This may reduce the impact of pruning only the top 10 attention heads.

\section{Glossary of Linguistic Terms}
\label{sec:appendix_glossary}

In this section, we include a glossary of some linguistic terms relevant to BLiMP that merit definition.

\paragraph{Anaphors and binding theory}
Binding theory conditions the distribution of nominals, particularly pronouns and anaphors, i.e., reflexive pronouns \citep{asudeh_binding_2006}.
The main constraint occurs with respect to a potential antecedent nominal that co-refers with the target nominal.
In the following discussion, this co-reference is signalled by indexing with $i$ and $j$.
Anaphors can only occur if there is a valid antecedent in the sentence, see the example in (\ref{ex:binding}) restated in (\ref{ex:binding_app}):

\begin{exe}
    \ex\label{ex:binding_app}
    Gina$_i$ explains Alan$_j$ fires himself$_j$.
\end{exe}

Pronouns can have a valid antecedent but not in the same clause as them:

\begin{exe}
    \ex Gina$_i$ explains Alan$_i$ fires her$_i$.
\end{exe}

\paragraph{Control and raising}
Both constructions involve a noun phrase (NP) in a main clause determining a covert reference of a subject of a subordinate clause.
The key difference is that under raising this main clause NP (in $M$) receives its semantic role from the verb of the subordinate clause `\textbf{expected}' (in $S$), see in (\ref{ex:raising}).
Under control, on the other hand, the NP receives its semantic role from the verb of the main clause `\textbf{persuaded}' (\ref{ex:control}).

\begin{exe}
    \ex\label{ex:raising}
    $[_M$ The teacher \textbf{expected} \underline{the students} $]_M$ $[_S$ to help the visitors $]_S$.
    \ex\label{ex:control}
    $[_M$ The teacher \textbf{persuaded} \underline{the students} $]_M$ $[_S$ to help the visitors $]_S$.
\end{exe}

In practice, however, both constructions involve a subordinate that is seemingly lacking a subject \citep{dubinsky_control_2006}.

\paragraph{Ellipsis}
Ellipsis refers to the "omission of linguistic material, structure, and sound" \citep{winkler_ellipsis_2006}.
Certainly ellipsis is not an unbounded phenomenon, and typically the omitted material stands in a co-reference relation with some overt element in the sentence.

\paragraph{Filler-gap dependencies}
Also called long-distance dependencies, `filler' refers to a fronted element and the `gap' is the position with which it is semantically or syntactically related \citep{falk_long-distance_2006}.
Examples for such dependencies involve \textit{wh} questions, exclamatives, topicalisation, cleft, and more, see the examples in (\ref{ex:filler-gap_app}).

\begin{exe}
    \ex\label{ex:filler-gap_app}
    \begin{xlist}
        \ex \underline{Which painting} does the artist believe the curator said the gallery owner hung $t$ on the wall?
        \ex \underline{What a painting} the artist believes the curator said the gallery owner hung $t$ on the wall!
        \ex \underline{This painting}, the artist believes the curator said the gallery owner hung $t$ on the wall.
        \ex \underline{It is the painting} that the artist believes the curator said the gallery owner hung $t$ on the wall.
    \end{xlist}
\end{exe}

\paragraph{Island effects}
Under certain (generative) linguistic analyses, various elements of a sentence materialise in a different location than where they were generated in \citep{carnie_island_2006}.
For instance, \textit{wh}-movement, i.e., the displacement of a question word is a common example for this, see (\ref{ex:wh}).

\begin{exe}
    \ex\label{ex:wh} 
    \begin{xlist}
        \ex Mary does love pineapples.
        \ex What does Mary love $t$?
    \end{xlist}
\end{exe}

The trace $t$ indicates where the question word \textit{What} originated from under the movement analysis.
Islands, on the other hand, are a collection of constraints on movement.
These typically contain restrictions on moving across various clauses.
There are many such constraints, but one example is the so-called Coordinate Structure Constraint that disallows extracting members of a conjunction, see the examples in (\ref{ex:csc}).

\begin{exe}
    \ex\label{ex:csc}
    \begin{xlist}
        \ex I have eaten the salad and the pizza.
        \ex *What have you eaten the salad and $t$?
        \ex *What have you eaten $t$ and the pizza?
    \end{xlist}
\end{exe}

\paragraph{NPI}
Negative Polarity Items (NPI) -- like \textit{any}, \textit{ever}, or \textit{yet} -- are words that appear only in a limited number of contexts.
These contexts are among all, the scope of negation, as complements of negative predicates, in comparative clauses, in questions, and in the scope of negative quantifiers and adverbs such as \textit{few/little}, \textit{rarely}, or \textit{only} \citep{hoeksema_polarity_2006}.

\end{document}